\documentclass{springer_template/svproc}

\usepackage{microtype}
\usepackage{graphicx}
\usepackage{subfigure}
\usepackage{booktabs}
\usepackage{wrapfig}
\usepackage{algorithm}
\usepackage{algpseudocode}
\usepackage{hyperref}

\usepackage{amsmath}
\usepackage{amssymb}
\usepackage{mathtools}

\usepackage[capitalize,noabbrev]{cleveref}

\usepackage[textsize=tiny]{todonotes}
\newcommand{\repeatthanks}{\textsuperscript{\thefootnote}}

\begin{document}

\mainmatter              
\title{Reinforcement Learning of Self Enhancing Camera Image and Signal Processing}
\titlerunning{Self Enhanced Camera ISP}  
%
\author{Chandrajit Bajaj\thanks{These authors contributed equally to this work}  \and Yunhao Yang\repeatthanks \and Yi Wang\repeatthanks}
\authorrunning{Chandrajit Bajaj, Yi Wang and Yunhao Yang} 
%
\tocauthor{Chandrajit Bajaj, Yi Wang and Yunhao Yang}
\institute{Department of Computer Science and Oden Institute of Engineering and Sciences, University of Texas, Austin, Texas, USA}

\maketitle              

\begin{abstract}
Current camera image and signal processing pipelines (ISPs), including deep-trained versions, tend to apply a single filter that is uniformly applied to the entire image. This is despite the fact that most acquired camera images have spatially heterogeneous artifacts. This spatial heterogeneity manifests itself across the image space as varied Moire ringing, motion-blur,  color-bleaching, or lens-based projection distortions. Moreover, combinations of these image artifacts can be present in small or large pixel neighborhoods, within an acquired image. Here, we present a deep reinforcement learning model that works in learned latent subspaces, and recursively improves camera image quality through a patch-based spatially adaptive  artifact filtering and image enhancement. Our \textit{Recursive Self Enhancement Reinforcement Learning}(RSE-RL) model views the identification and correction of artifacts as a recursive self-learning and self-improvement exercise and consists of two major sub-modules: (i) The latent feature sub-space clustering/grouping obtained through variational auto-encoders enabling rapid identification of the correspondence and  discrepancy between noisy and clean image patches. (ii) The adaptive learned transformation is controlled by a soft actor-critic agent that progressively filters and enhances the noisy patches using its closest feature distance  neighbors of clean patches. Artificial artifacts that may be introduced in a patch-based ISP, are also removed through a reward-based de-blocking recovery and image enhancement.  We demonstrate the  self-improvement feature of our model by recursively training and testing on images, wherein the enhanced images resulting from each epoch provide a natural data augmentation and robustness to the RSE-RL training-filtering pipeline. Our method shows advantage for heterogeneous noise and artifact removal. Our code is publicly available at \url{https://github.com/CVC-Lab/rse_rl.git}.
\end{abstract}
\section{Introduction}
\label{sec:intro}

Image Processing Pipelines for cameras, known as camera ISPs, have been a hot topic in past decades. For instance, image denoising has been investigated due to the vast demand for noise pattern removal. In the early era, traditional methods would explore the case where they apply Gaussian blurring, TV regularization \cite{rudin1992nonlinear}, or a coefficient transform in Fourier domain \cite{simoncelli1996noise} to remove high-frequency signals, commonly identified as noise signals. However, it is the idea of non-local mean denoising from \cite{buades2005non} that truly starts a heterogeneous noise removal. The non-local mean method is built upon self-similarity and redundant information over realistic images, without a uniform noise pattern assumption, which is proved to lead a performance boost. Later on, BM3D \cite{dabov2007image}, as a non-local method, explore the sample idea and exploits the sparsity further. There are discussions with respect to subdivision of image domains \cite{ghimpecteanu2016local} as well. Nowadays, with the advent of deep learning architectures, more and more works are using Convolutional Neural networks (CNN) or even Generative Adversarial networks (GAN) that beats most of the classical, sophisticated methods \cite{zhang2017beyond,guo2019toward,xu2020noisy,moran2020noisier2noise,hou2020nlh}. Nevertheless, as noise in large-scale realistic photos is obliged to an unknown form, most of the prior works, including those with a deep learning denoising model, depend on an assumption that noise comes from easily synthesized additional signals. The most common example is the AWGN noise model. AWGN noise model cannot effectively assist either a machine learning model or a deep learning model to a better improvement in real-photo noise/artifact removal, as discussed in earlier benchmarks \cite{plotz2017benchmarking}. How to fill in the gap between the synthesized noise model and real-world noisy images remains an open question.

In recent years, more self-supervised learning or unsupervised image denoising models have aroused. These models aim at finding an agnostic noise modeling under real photographs, rather than synthesizing its noisy Version from instrument error or Gaussian/Poisson noise distortions. Among all works, N2V\cite{krull2019noise2void} and N2N\cite{moran2020noisier2noise} are typical examples of such learning schemes, and later works \cite{xie2020noise2same,jang2021c2n} explores a more generative, latent space representation of noise model equipped with more flexibility of learning dependently from image signals. Even though self-redundancy and realistic noise generation are partly solved in these proposed models, these models rely on the assumption that there is a single image artifact per RAW image that needs to be diagnosed and filtered. Heterogeneous camera image artifacts, especially mixtures, are insensitive to pixel locations and thus it is hard to disentangle them from one pass. For a camera ISP, however, this assumption fails to capture spatial heterogeneous acquired corruption caused by a mismatch of acquisition settings and environmental lighting in realistic scenes. Moreover, one can hardly tell if the artifacts come from sensor limitations, environmental changes or post-processing such as lossy compression. Such varied real-world image processing issues motivate us to adapt a recursive self-improving machine learning approach for the next generation of camera ISPs. Although reinforcement learning-based image improvement pipeline was invented in parallel \cite{hu2018exposure,yu2018crafting,furuta2019pixelrl,zhang2021r3l}, none of them take care of spatial heterogeneity compared with the unsupervised or self-supervised scheme. It is worthwhile to combine the idea of recursive improvement together with a spatial subdivision,
and, if possible, control and understand patch-based latent space representation of realistic photos.

In this paper, we present a deep reinforcement learning model that recursively improves camera image quality through a patch-based spatially adaptive artifact filtering. Our Recursive Self Enhancement Reinforcement Learning(RSE-RL) model is introduced in Section \ref{sec:main:algorithm}. The main contribution of RSE-RL model is a two-fold patch-based denosing filtering. On the one hand, we learn latent space feature disentanglement by projecting the encoding of corrupted patches into subspaces. With the discrepancy being identified through transformation from subspaces, one can assign the decoding filters, which is a neural network, to recover the clean image. On the other hand, we iteratively refine our denoising result via changing the transformation through actions, optimizing the performance per test input.   We demonstrate, in Section \ref{sec:experiment}, the performance of our RSE-RL model including the self-improvement features by recursively training and testing on images, wherein the enhanced images resulting from each epoch provide a natural data augmentation and robustness to the RSE-RL training- filtering pipeline.

\section{Related Work}

\subsection{Image Denoising}

\paragraph{Generative Noise Patterns}

The trend of current image denoising research focuses more on model-free noise generators, no matter globally or locally. \cite{liu2021invertible} adapts a Wavelet Transform as a separator for high-frequency signals and then applies the design of the invertible blocks to recover and refine low-resolution image signals. \cite{xia2021lowlight} discusses the contrastive learning under flashing/no-flashing pairs to cancel out noise under low-light conditions. \cite{ren2021adaptive} propose the consistency prior which measures the correlation and tries to linearize the image noise model.  \cite{khademi2021pgnoise} extend the blindspot model for self-supervised de-noising to handle Poisson-Gaussian noise specifically. Noise2same \cite{xie2020noise2same}, as a recent attempt at self-supervised learning, takes both the full noisy image and the masked image as inputs and produces two outputs. The reconstruction loss is computed between the full noisy image and its corresponding output. The invariance loss is computed between the two outputs to align with a standard denoising filter. Among all these works, the generative model of noise is assumed as a global modification. 

\paragraph{Latent Subspace Denoising}

One of our major proposed contributions is to learn a latent encoding for maximizing the usage of self-similarity in natural images via probing sub-spaces. Earlier work, such as \cite{strela2000image} propose a Gaussian mixture model for global noise. Inspired from latent space clustering using VAE \cite{dilokthanakul2016deep}, \cite{yang2021learning} introduces a latent mixture subspace and optimizes under multiple distinguished denoising filters. The variational inference under the latent space disentanglement yields a better separability and approximation power towards a non-uniform, signal-dependent noise removal algorithm. Compared with \cite{liu2021gmm}, where they study the influence of Gaussian noise on  parameters of Gaussian Mixture Model. The result can adapt to any noise prior. However, for unseen data, its generalizability remains a question and the real noisy image might fall into a simple filter's scope.  Another attempt of latent space representation of noise modeling is NBNet
\cite{cheng2021nbnet}, which learns latent space noise basis under projection with principal component analysis. In addition, for patch-based subspace noise learning, \cite{jang2021c2n} concatenate the whole image with its noise level map taken as the encoder input. Given a query patch, similar context patches will be found by cross-patch sampling. It comprises two branches:  the correlation of the inner patch grasp for the local feature while a cross-patch GCN extracts patch-wise consistency. Then the model aggregates the local and non-local features as the input of the decoder and applies residual blocks for fine-detailed denoise image output. PS-VAE\cite{yang2021learning} proposed to impose Gaussian Mixture Model in the latent space encoding for patch-based denoising. A soft-labeling was introduced to assign different filters for patches. The model has the potential of denoising heterogeneously. Nevertheless, the prior of noise is intractable in a real photo, leading to a overly strong assumption which is difficult to generalize.

{
\subsection{Camera ISP meets Deep Learning Model}
 Deep learning approaches have progressively replaced the image and signal processing applied in conventional computational photography tasks. For instance, with low-level details and hierarchical structures of neural network implemented, one could achieve superior performance for image deblurring (e.g. \cite{abdelhamed2019ntire,xu2019learning,tian2020deep,valsesia2020deep})  and deblurring (e.g. \cite{kupyn2019deblurgan,suin2020spatially}) tasks. There are numerous works related to camera ISP as well, where different types of articulated modelings are applied to different vision tasks. Among all of them, some related works such as color demosaicing(e.g. \cite{khashabi2014joint}), image denoising(e.g. \cite{tian2020deep}), auto white balance correction(e.g. \cite{afifi2020deep}), removing lossy compression artifacts (e.g. \cite{galteri2017deep}), \emph{etc.}, have been separately discussed under deep learning settings.

Within the scope of image denoising, CycleISP \cite{zamir2020cycleisp} develops a generative model to generate synthesized realistic image data that is both forward and reversed. Moreover, solving multi-task image enhancement is possible: \cite{schwartz2018deepisp} is an early work that focuses on jointly solving demosaicing and denoising problems by applying deep learning schemes from the source sampling of sensors. A recent work \cite{ignatov2020replacing} presented a single deep learning model containing 5 parallel learning levels to replace the entire camera ISP pipeline, referred to as PyNET. The input RAW image from a cellphone is aligned with a DSLR camera output as the supervised training data and PyNET outputs a visually high-quality sRGB image. To solve noise pattern generation issues, \cite{cao2021pseudo} focus on synthesizing noise pattern which is dependent of RAW image signals, and iteratively update the denoised image from an unpaired set of noisy and clean images to adapt and finetune the pre-trained denoiser.
}

\subsection{Reinforcement Learning based Image Enhancement}

Reinforcement Learning methods have been applied in different low-level vision tasks, especially in image processing. \cite{furuta2019pixelrl} has proposed a pixel-wise A3C (Asynchronous Advantage Actor Critic\cite{mnih2016asynchronous}) scheme to view the entire image domain as a multi-agent system, but provide their own filtering action and reward map convolution to efficiently compute mean reward. This early system has enlightened the later research to formulate the image processing task as multi-agent, or multi-task learning problem, which could be controlled via reinforcement learning. A global image enhancement approach was introduced in \cite{hu2018exposure}, where different filters are applied over the RAW image with a decision-making procedure, i.e. a policy gradient descent. The input of the network is a low-res, fixed 64 by 64 patches where the patch is used to determine which filter and the weight of filters applying over the high-res image. \cite{yu2018crafting} extends the idea of filter decision but the decision-making is based on Long-Short Term Memory(LSTM) rather than a conventional CNN as stated in \cite{hu2018exposure}. 

\begin{figure}[t]
\begin{center}
    \includegraphics[width=0.9\linewidth]{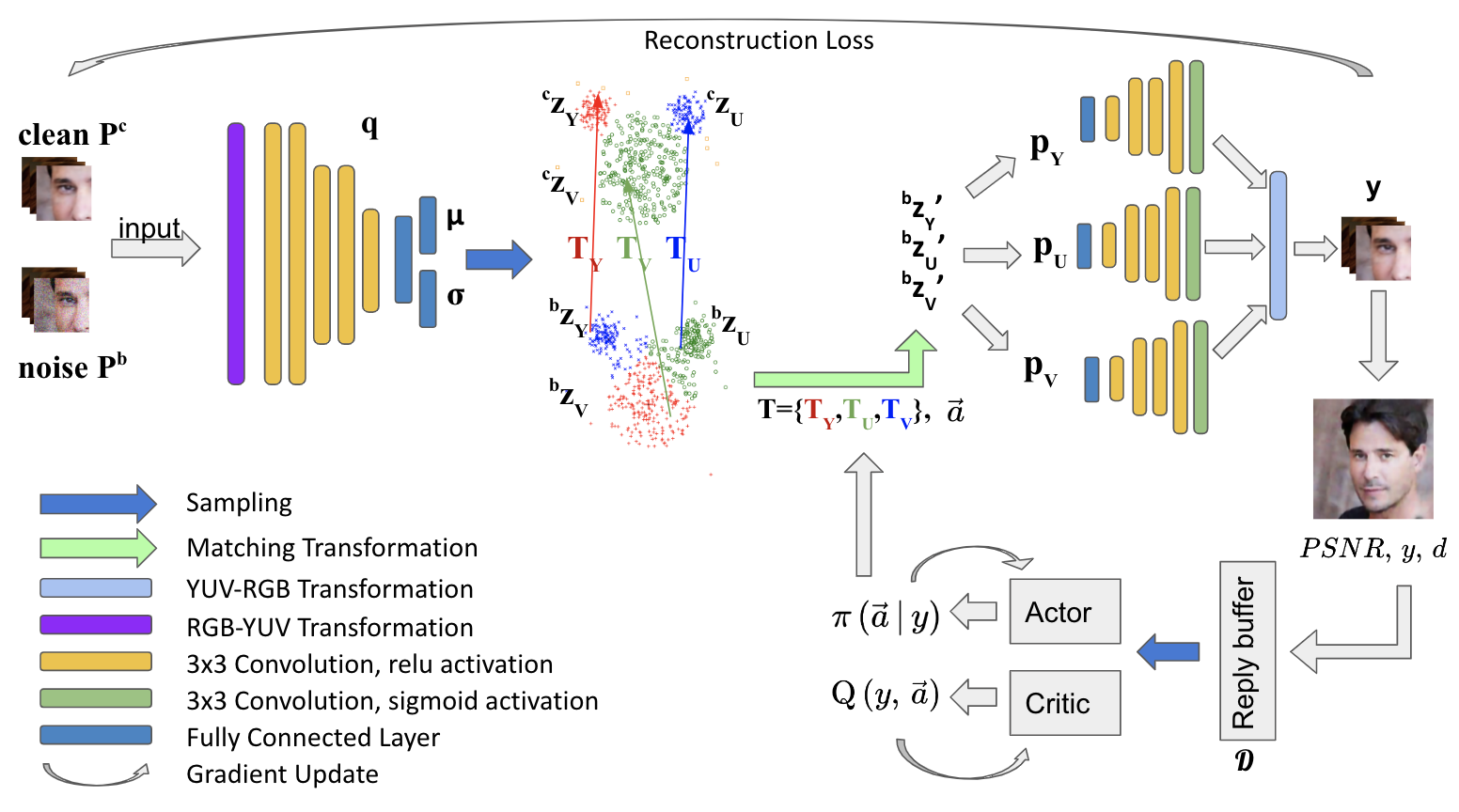}
\end{center}
\caption{The overall pipeline of our RSE-RL: For each captured (and training) image, we split the image into local patches and feed every patch as a stack into the encoding network.
We divide the latent space into three subspaces $Z_y$, $Z_u$, and $Z_v$ which preserve the YUV features of the patches.
We apply the encoder to the clean and noisy patches and  project them as well onto the  same three latent subspaces.
Then, we learn a set of transformations $T$ that transforms the latent representation of the noisy patches to a corresponding representation of the clean patches, in all three subspaces. We send individually sampled transformed noisy representations to the decoders to reconstruct the image. Note that we have three decoders to reconstruct YUV features from the three subspaces. Then we merge and reconvert from the YUV features to reconstruct the final RGB image.
Once we reconstruct the images, we train a soft-actor-critic reinforcement learning algorithm to further maximize the image PSNR.
The RL algorithm uses s 1-norm distance between a target PSNR and the actual PSNR as the reward to fine-tune the trainable weights in the set of transformations $T$. Application of the RL  self-enhancing image denoising network. More details are presented  below and in the Appendix.}
\label{fig:struct:final}
\end{figure}

\section{Recursive Self-Enhancing Camera ISP}

\label{sec:main:algorithm}
In this section, we describe the architecture of our reinforcement learning based camera ISP model that recursively improves and enhances acquired images by filtering heterogeneous image artifacts in a spatially adaptive manner.

We target the problem of resolving image artifacts, specifically on image denoising, but our approach can be extended to solve other comprehensive image processing tasks, such as generating RAW to sRGB images \cite{ignatov2020replacing,zhang2021learning}. We assume the observed image $I_{obs}$ is obtained via the following mixture model:
\begin{equation}
    f(I_{obs}) = I_{gt} + \sum_{s=1}^S \Sigma_s \odot M_s.
    \label{eq:model:mixture:noise}
\end{equation}

The function $f$ is an identity function performing sRGB to sRGB image denoising tasks. The noise $\Sigma_s$ and the mask matrix $M_s$ are independent and blind to the model ($\odot$ refers to the element-wise product). Moreover, we do not rely on the underlying distribution $\Sigma_s$. Except for the synthesized dataset, we cannot obtain the accurate number of artifact types $S$, and $S$ is a hyper-parameter in most scenarios. We would like to present and discuss how to disentangle and filter from (\ref{eq:model:mixture:noise}) using our RSE-RL. We will first train distinct filters in separated latent subspace with supervision. Then we impose self-enhancement for testing data, by learning per-image latent transformation under a policy learning scheme. 

\paragraph{Overall Pipeline of RSE-RL}
Our recursive self-improving camera ISP in Figure \ref{fig:struct:final} consists of a multiple latent subspace variational autoencoder developed from \cite{yang2021learning}. For every input image $I_{obs}\in\mathbb{R}^{H\times W\times C}$, we first divide the input image into $D$ by $D$ patches $P\in\mathbb{R}^{D\times D \times C}$ with overlaps allowed. For every $D$ by $D$ patch $P$ in the image, we denote its location mask $H_n$ in the original image domain. $P$ relates to the observed image in the form $P\overset{def}{=} I_{obs}\odot H_n$. Even though we do not estimate the magnitude of the artifacts, we approximate it  within a patch with single dominant model, namely:
\begin{equation}
    f(P) = f(I_{obs} \odot H_n) =  I_{gt}\odot H_n + \sum_{s=1}^S  \Sigma_{s} \odot M_{s} \odot H_n , \approx (I_{gt} + \Sigma_i \odot M_i) \odot H_n.
\label{eq:model:mixture:patch}
\end{equation}

Our RSE-RL network initially  learns from clean image patches $P^c=I_{gt}\odot H_n$ and observed noisy patches $P^b$. We feed both  clean patches $P^c$ and noisy patches $P^b$ into the variational encoder $q(P|\vec{\theta})$ parametererized by $\vec{\theta}$. The encoder projects the patches into three latent subspaces $Z_y$, $Z_u$, and $Z_v$ and we draw samples in three latent spaces via:
\begin{equation}
    \vec{z}_s \sim \mathcal{N}(q_s^{\mu}(P|\vec{\theta}),diag [(q_s^{\sigma}(P|\vec{\theta}))^2]),\quad s\in \{y,u,v\}. 
\end{equation}
We denote $\{\vec{z}_{y}^b,\vec{z}_{u}^b, \vec{z}_{v}^b\}$ and $\{\vec{z}_{y}^c,\vec{z}_{u}^c, \vec{z}_{v}^c\}$ as the latent encodings (samples) for noisy patches $P^b$ and clean patches $P^c$ respectively. To obtain latent subspaces which learn to identify imminence and chrominance components, we apply an RGB to YUV transformation per pixel for each input patches $P^b$ and $P^c$ as a pre-processing layer in encoder $q$. We state the RGB-YUV transformation matrix in the Appendix.

Within the three subspaces $Z_y$, $Z_u$, and $Z_v$, we construct a set of three transformations $T = \{T_y, T_u, T_v\}$. We train the transformations so that the transformed latent encoding approaches the clean patch latent encoding: $T(\vec{z}^b) \approx \vec{z}^c$. Then, we set up three decoders $P_y:= p_y(T_y(\vec{z}^b_y)|\vec{\psi}_y), P_u:=p_u(T_u(\vec{z}^b_u)|\vec{\psi}_u), P_v:=p_v(T_v(\vec{z}^b_v)|\vec{\psi}_v)$ with parameters $\vec{\psi}_s, s \in \{y, u, v\}$ respectively for reconstructing YUV channels from three disentangled latent subspaces. After the YUV features are constructed, we transform the patches' YUV features into RGB patches under a patch transformation $\mathit{PT}$, which is a pixel wise YUV to RGB mapping:
\begin{equation}
    \mathit{PT}(P_y, P_u, P_v) = vec_{D\times D \times C}^{-1} \left( A \begin{bmatrix}vec(P_y),vec(P_u),vec(P_v)\end{bmatrix}^T\right).
\end{equation}

Here the $vec_{D\times D\times C}^{-1}$ is the reshaping operator that reshapes the patch back to $D$ by $D$ by $C$. The YUV-RGB patch transformation matrix $A$ is the inverse of RGB-YUV linear transformation matrix, whose form is given in the Appendix. We reconstruct each patch independently and merge them after the decoding procedure. The output is the combined reconstructed patches and image that approximate $P^{c}$.

\paragraph{Loss Function} 

In practical implementation, our training process optimizes different loss functions. For training the network, our training loss $\mathcal{L}_{vae}$ consists of three parts: the evidence lower-bound(ELBO) \cite{Kingma2014}, including the data fitting term and the KL loss, the transformation constraint loss, and the regularization term. The gradient computed from $\mathcal{L}_{vae}$ is used to update all the parameters in the network, including the encoder, decoders, and transformations.
\begin{equation}
        \mathcal{L}_{vae}= \mathcal{L}_{MSE}+ \mathcal{L}_{KL}+ \mathcal{L}_{reg} +\mathcal{L}_{tran},
    \label{eq:loss:all}
\end{equation}
where
\begin{align}
    \mathcal{L}_{MSE} =&  \| \mathit{PT}(P_y, P_u, P_v)- P^c \|_F^2 +\| \mathit{PT}(\tilde P_y, \tilde P_u, \tilde P_v) - P^c \|_F^2 , \\
    \tilde P_s :=& p_s(\vec{z}_s^c|\vec{\psi}_s), \quad s \in \{y, u, v\}, \\
      \mathcal{L}_{KL} =& - \frac{1}{2} \sum_{s\in \{y, u, v\}}\sum_j (1+ 2\log q_{s,j}^{\sigma}(P|\vec{\theta}) - (q_{s,j}^{\sigma}(P|\vec{\theta}))^2 - (q_{s,j}^{\mu}(P|\vec{\theta}))^2), \label{eq:loss:enc:KL} \\
    \mathcal{L}_{reg} =& \lambda_{reg}(\|\vec{\theta}\|_2^2 + \|\vec{\psi}\|_2^2),\label{eq:loss:enc:reg} \\
    \mathcal{L}_{tran}= & \|\vec{z}_y^c -T_y(\vec{z}_y^b)\|_2^2 + \|\vec{z}_u^c -T_u(\vec{z}_u^b)\|_2^2 + \|\vec{z}_v^c -T_v(\vec{z}_v^b)\|_2^2.
    \label{eq:loss:tran}
\end{align}
On each latent subspace, we compute a loss $\mathcal{L}_{tran}$ from the noisy patch projection and clean patch projection and update the corresponding transformation accordingly.

\paragraph{Patch Assembly and Block Artifacts}

Upon getting the reconstructed blocks from the decoders, we merge these blocks back into the original image. If one naively concatenates two adjacent patches without any tolerance of the overlaps generated, one could observe block artifacts if these two patches pass different decoders. We apply post-processing to remove block artifacts generated from concatenation. We average the pixel output for overlapping regions based on the distance between the pixel and the overlapped patch centers via linear interpolation.

\paragraph{Reinforcement Learning}

New images acquired in the test or field operational stage, are reutilized to retrain the model. Since her we do not have access to the corresponding clean target patches, the trained transformations are fine tuned based on a reward function comprising of a target PSNR larger than the acquired image PSNR. 

We use a Soft Actor-Critic (SAC) reinforcement learning algorithm \cite{haarnoja2018soft,haarnoja2018softb} as the self-enhancing mechanism. In our setup, the SAC algorithm learns to iteratively enhance from the current state of image patches $P$ by updating the latent space transformation $T$ using an weight vector $\vec{a}$. $\vec{a}$ can be viewed as a vectorial action, comprising of three component actions $\vec{a}_y,\vec{a}_u$ and $\vec{a}_v$, in the YUV latent subspaces. If we parameterize our transformation network $T_s,s\in\{u,v,w\}$ with parameter $\xi_s$, the weight vector update action can be expressed as an element-wise multiplication \(
    \vec{\xi}^{\prime}_s = a_s \odot \vec{\xi}_s.
\)
Here $\vec{\xi}^{\prime}_s$ is the updated weight, and $\vec{\xi}_s$ is the previous weight. The algorithm continuously updates these trainable weights to enhance the transformation from noisy to clean patch representations, consequently improving the final enhanced and filtered result.

We briefly explain SAC here and how it applies to our problem. SAC is an off-policy algorithm with entropy  regularization. SAC trains an RL model to maximize the trade-off between expected return and entropy, a measure of randomness in the policy. SAC can prevent the policy from prematurely converging to a bad local optimum. In SAC, an entropy bonus is reflected in $Q^{\pi}$:
\(
    Q^{\pi}(P,a)\approx r(P,a) + \gamma(Q^{\pi}(P',\tilde{a}') - \alpha \log\pi(\tilde{a}'|P')),\ \tilde{a}'\sim \pi(\cdot|P'),
\)
where $P$ is an input patch and $P$' is the resulting patch when the transformation in the network has been updated using $\tilde{a}'$, $\alpha > 0$ is the trade-off coefficient, $r(P,a,P')$ is the reward function and $\log\pi(\tilde{a}'|P')$ is the defined entropy.

SAC simultaneously learns a policy $\pi$ and two Q functions $Q_{\phi_1}, Q_{\phi_2}$, with $\phi_1,\phi_2$ the network parameters. The actor-network learns the $Q$ functions that minimizes a Mean-Squared Bellman Error(MSBE) $L(\phi_{i},D)$, and SAC sets up the MSBE loss for each of the two $Q$-functions:
\begin{equation}
   \mathcal{L}(\phi_j,D) = E\left[((Q_{\phi_j}(P,a)-f(r,P',d))^{2}\right], \quad(P,a,r,P',d)\sim D
   \label{eq:sac:L}
\end{equation}
where $d$ is the done signal to set a terminating state, and $f$ is given by
\begin{equation}
    f(r,P',d) = r + \gamma (1-d)(\min_{j=1,2} (Q_{\phi_j}(P',\tilde{a}')-\alpha log\,\pi_{\theta }(\tilde{a}'|P'))),\quad \tilde {a}'\sim \pi(\cdot |P')
    \label{eq:sac:f}
\end{equation}

In the SAC algorithm, we consider patches from an image as state observations. The reward $r$ is given by computing the L1 distance between the actual patch PSNR and a target patch PSNR for each of a  mini-batch of samples. Namely,
\begin{equation}
\label{eq:reward}
    r(P,a,P') = b \left(PSNR(PT(P_y, P_u, P_v), P') - PSNR_t \right) + c,
\end{equation}
where $b$ and $c$ are constants for adjusting the reward scale. $PSNR$ is the quality metric, $N$ is the total number of patches, and $PSNR_t$ is the target PSNR of the entire patch-assembled image. The SAC trains an RL model to enhance the transformation and reduce the blocking artifacts. Thus, the optimized RL model can enhance images PSNR. We present more details and hyper-parameter settings in the Appendix. 

The critic network learns the policy $\pi_{\omega}$ by maximizing a value function $V^{\pi}(I)$ to evaluate the current policy and computes gradients for the actor to find the optimal policy updating $a$ accordingly via sampling from $\pi_{\omega}$.
\begin{equation}
\begin{split}
        \mathcal V^{\pi}(P) &= E_{a\sim \pi_{\omega}}[Q^{\pi}(P,a)]+\alpha H(\pi_{\omega}(\cdot |P))\\&=E_{a\sim \pi_{\omega}}[Q^{\pi}(P,a)-\alpha log\,\pi_{\omega}(\cdot |P)].
\end{split}
\label{eq:sac:v}
\end{equation}
Here $Q^{\pi}$ takes the minimum value between $Q_{\phi_1}$ and $Q_{\phi_2}$. We refer to  readers the overall algorithm of our RSE-RL pipeline stated in Algorithm \ref{algo:main}.

\begin{algorithm}[H]
\caption{RSE-RL algorithm}
\label{algo:main}
\begin{algorithmic}
\State \textbf{Input}: Training patches $\{P_n^{b},P_n^{c}\}_{n=1}^N$, training steps MAX\_TRAIN\_ITER, minibatch size $M$, target noisy image $I_{test}$, gradient descent step size $\eta$, target PSNR $PSNR_t$.
\State \textbf{Output}: Trained VAE with three distinct decoders, denoised image $\tilde{I}_{test}$.
\State \% Training Pipeline
\For{$i=1$ to MAX\_TRAIN\_ITER}
\State Sample a minibatch $\{P_m^{b},P_m^{c}\}_{m=1}^M$ from $\{P_n^{b},P_n^{c}\}_{n=1}^N$.
\State Update network parameters via gradient descent of \eqref{eq:loss:all} using $\{P_m^{b},P_m^{c}\}_{m=1}^M$.
\EndFor
\State \% Self-Enhancement Pipeline, which can be a decoupled procedure from training.
\State Collect patches $\{P_n\}$ from $I_{test}$.
\While{$PSNR < PSNR_t$}
\State Sample $P\sim \{P_n\}$, the observed state and $a \sim \pi_{\omega}(\cdot|P)$, the action.
\State $P'\leftarrow \mathit{PT}(P_y, P_u, P_v)$.
\State Compute $PSNR$ and $r(P,a,P')$
\State Set done signal $d \leftarrow (PSNR < PSNR_t)$ AND the trajectory is less then $40$ steps.
\State Update replay buffer $D\leftarrow D \cup \{P,a,P',r,d\}$.
\For{$i=1$ to MAX\_EXPLORE\_ITER}
    \State Sample a minibatch $\{P_m,a_m,P_m',r_m,d_m\}_{m=1}^M$ from D.
    \State Compute $f(r_m,P_m',d_m)$ defined in \eqref{eq:sac:f}.
    \State $\phi_j \leftarrow \phi_j - \frac{1}{M} \sum_{m=1}^M \eta \nabla_{\phi_j} \mathcal{L}(P_m,a_m,P_m')$ with $\mathcal{L}$ defined in \eqref{eq:sac:L}.
    \State $\omega \leftarrow \omega -  \frac{1}{M} \sum_{m=1}^M\eta \nabla_{\omega}\mathcal V^{\pi}(P_m)$ with $V^{\pi}$ defined in \eqref{eq:sac:v}.
\EndFor
\State Assemble $\tilde{I}_{test}$ from patches by applying learnt policy and the training networks 
\State \textbf{Return}: $\tilde{I}_{test}$.
\EndWhile
\end{algorithmic}
\end{algorithm}

\section{Experiments}
\label{sec:experiment}
In the experiment section, we validate our RSE-RL algorithm on two datasets: Synthesized Noisy CelebFaces Attributes (CelebA) Dataset\cite{liu2015faceattributes} and Smartphone Image Denoising Dataset (SIDD)\cite{abdelhamed2018high}. CelebA dataset consists of a set of faces of celebrities, and we apply Gaussian noise to each image in the dataset to form a synthesized noisy dataset. SIDD dataset consists of images with real artifacts generated by smartphone cameras under various light conditions. We apply our RSE-RL architecture to remove the artifacts in the image and expect that our architecture outperforms other baseline, denoising models. A baseline model is an ordinary variational autoencoder trained under identical settings with our RSE-RL; we denote it as \textit{Single Decoder VAE}.
\subsection{CelebA Synthesized Artifacts Denoising}
\label{sec:celeba}
\paragraph{Dataset Construction}
We collect images from the CelebA\_HQ/256 dataset, where the size of each image is $256 \times 256$ pixels. We apply uniform Gaussian noise to each image in the CelebA dataset (using OpenCV\cite{opencv_library}) to form a synthesized noisy image corresponding to the original clean image. We consider the clean image and its noisy image as an image pair.
We divide the image pairs into a training set and a validation set.
The training dataset consists of 2,250 image pairs, and the validation dataset has 11,250 image pairs. 
We divide each image into $16 \times 16$ pixels patches, with 4 pixels overlapping with the surrounding patches, and then feed these patches into our network.
\begin{figure}[t]
\begin{center}
    \includegraphics[width=0.9\linewidth]{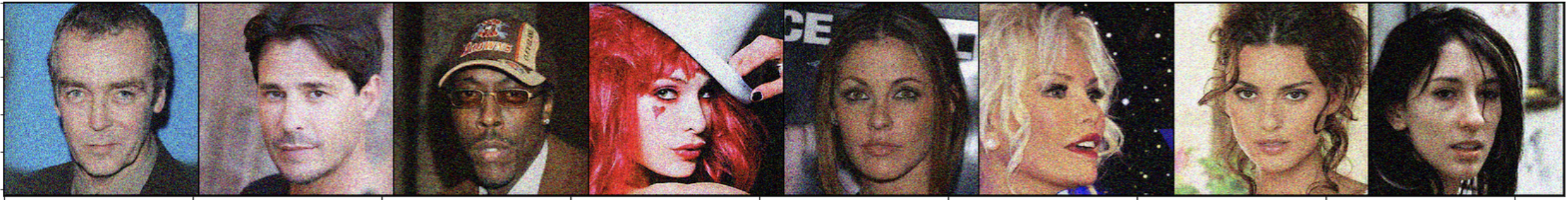}
    \includegraphics[width=0.9\linewidth]{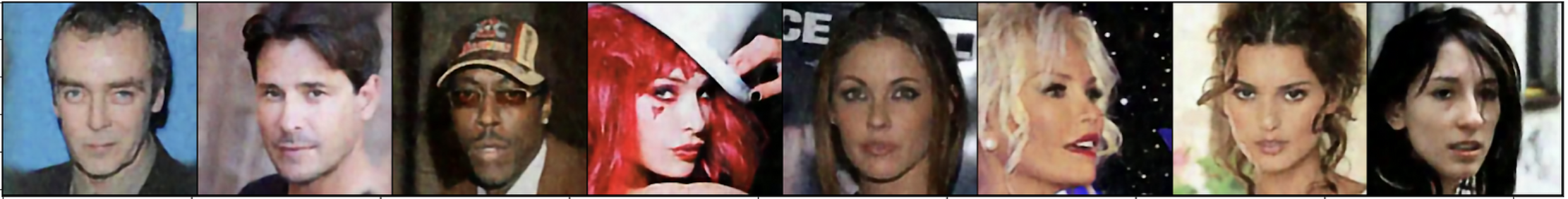}
\end{center}
\caption{CelebA Denoising Results: images on the top row are images that contain the synthesized artifacts (Gaussian Noise). The images on the bottom row are the denoising result from our RSE-RL.}
\label{fig:celeba:results}
\end{figure}
\paragraph{Experimental Setup}
We test the denoising performance of our RSE-RL over a synthesized noisy CelebA dataset. 
Our encoder projects the patch-based images into the latent space using 5 convolutional layers and 2 fully-connected layers in a subsequent order. The decoder has an inverse structure to the encoder. It has $2$ upsampling layers and 5 transposed convolutional layers. 
We implement the networks in Keras\cite{chollet2015keras} and Tensorflow\cite{tensorflow2015-whitepaper} and use a single 12GB NVIDIA Tesla K80 GPU for training and testing the synthesized noisy CelebA dataset.
We set the batch size to 128, epoch number to 50, regularization coeficient $\lambda_{reg} = 0.01$, optimizer to Adam \cite{Kingma2015adam} with $\beta_1 = 0.9$ and $\beta_2 = 0.999$. The learning rate is set by an exponential decay learning rate scheduler with an initial rate of 0.001, decay factor of 0.95, and decay step 1000.

\begin{figure}[t]
\begin{center}
    \includegraphics[width=0.5\linewidth]{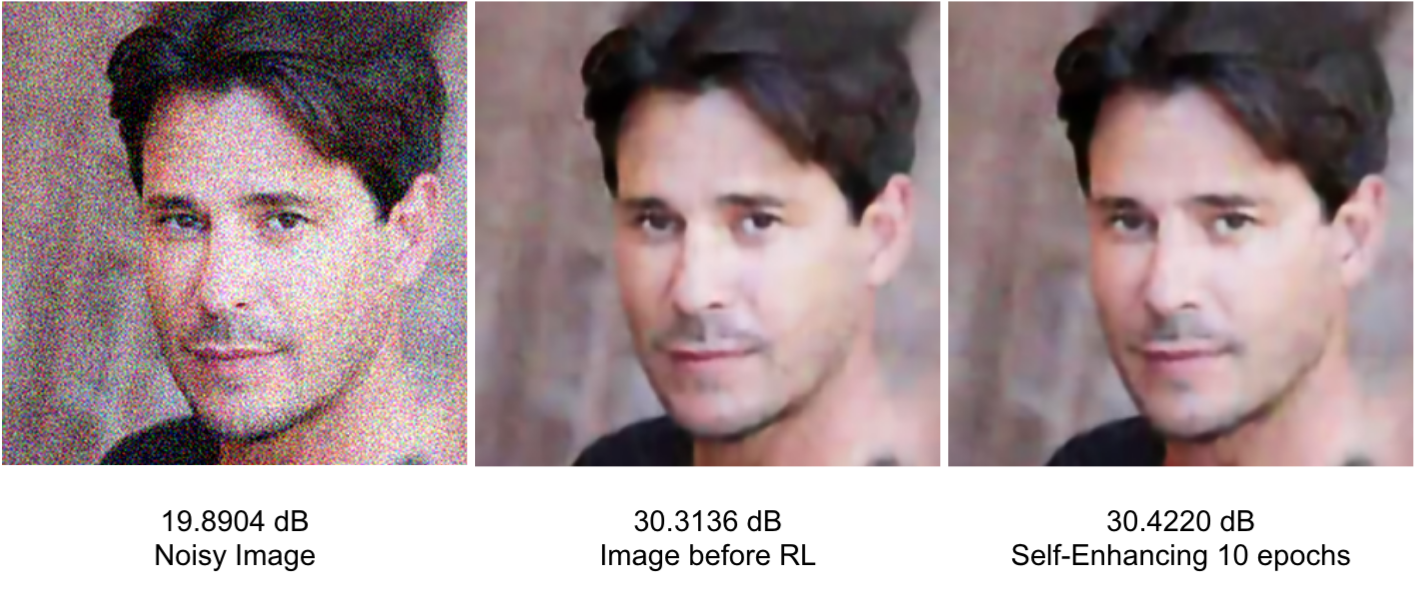}
    \includegraphics[width=0.5\linewidth]{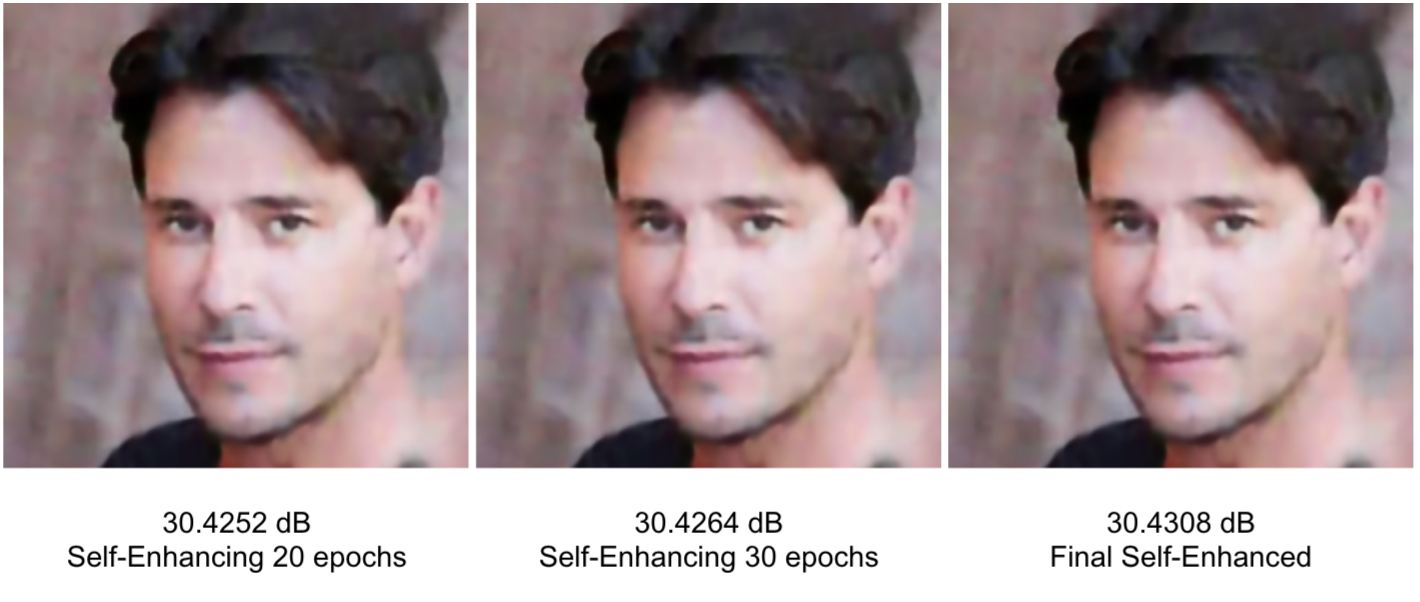}
    \includegraphics[width=0.5\linewidth]{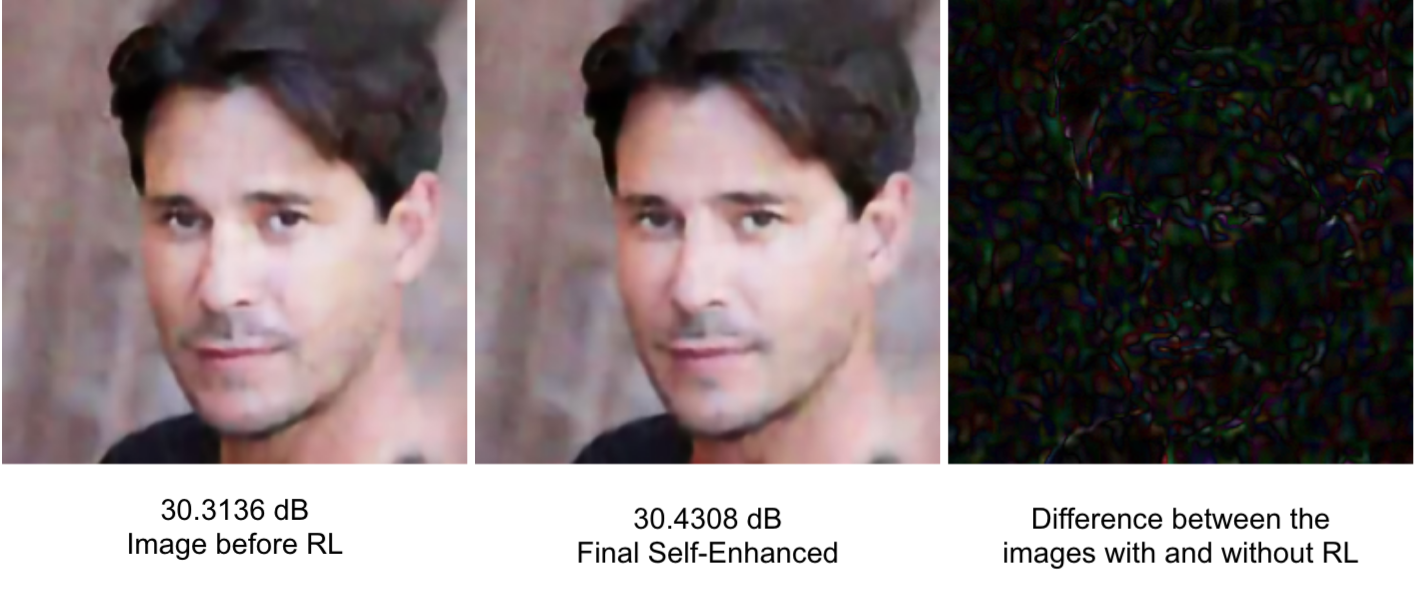}
\end{center}
\caption{Recursive Self-enhancing RL Visualization: the figure shows how the test images are recursively enhanced in a 50-epoch RL training. We observe a performance boost when iterating the weights of decomposed transformations under three latent subspaces. With the reinforcement learning agent, the network converges to a better result compared to the case where only a solo VAE framework can achieve. The last line also specifies the difference between the starting image we fed into the RL agent and the final result after recursive learning.}
\label{fig:recursive}
\end{figure}

\begin{figure}[!htbp]
\begin{center}
    \includegraphics[width=0.2\linewidth]{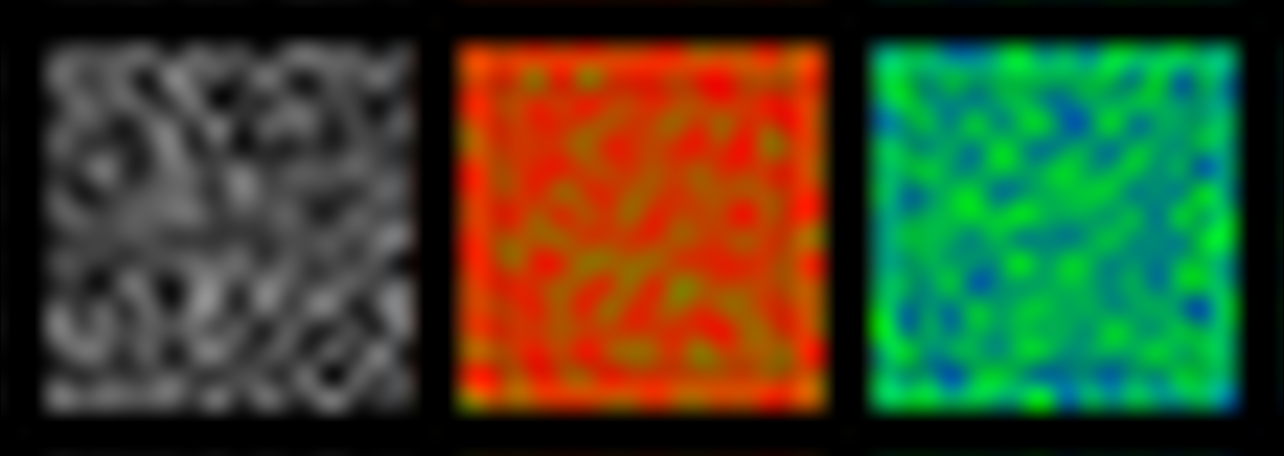}
    \includegraphics[width=0.2\linewidth]{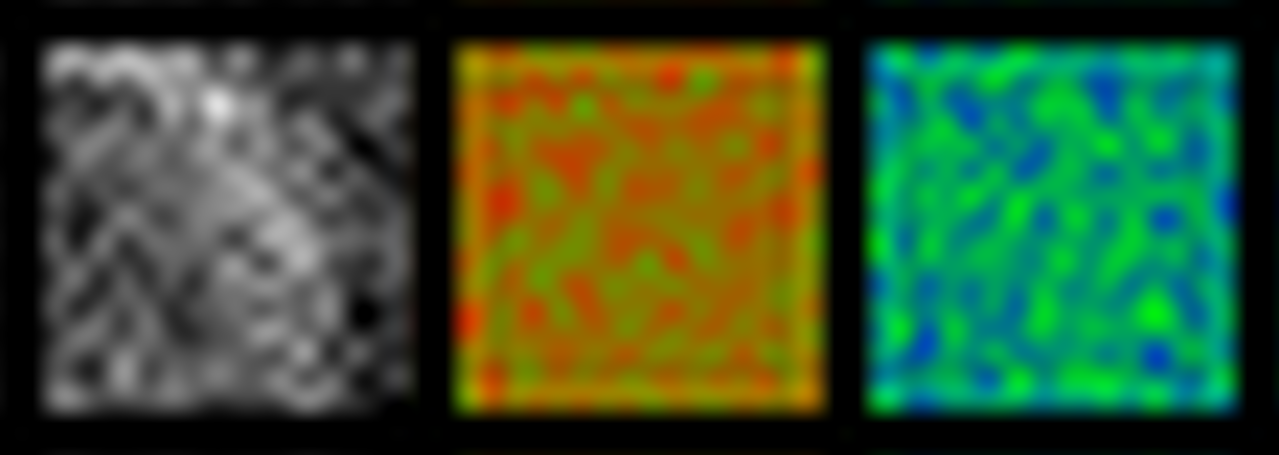}
    \includegraphics[width=0.2\linewidth]{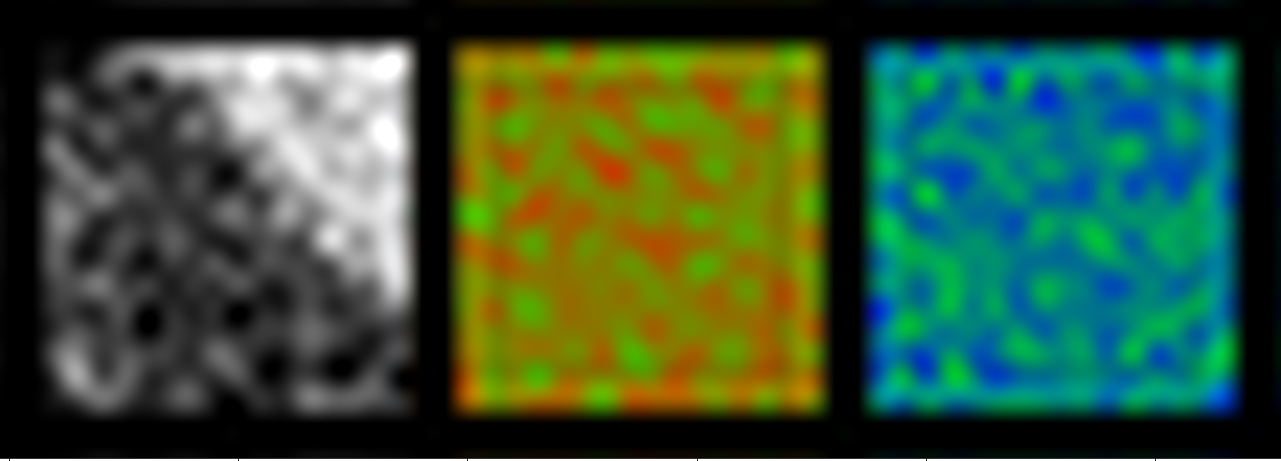}
    \includegraphics[width=0.2\linewidth]{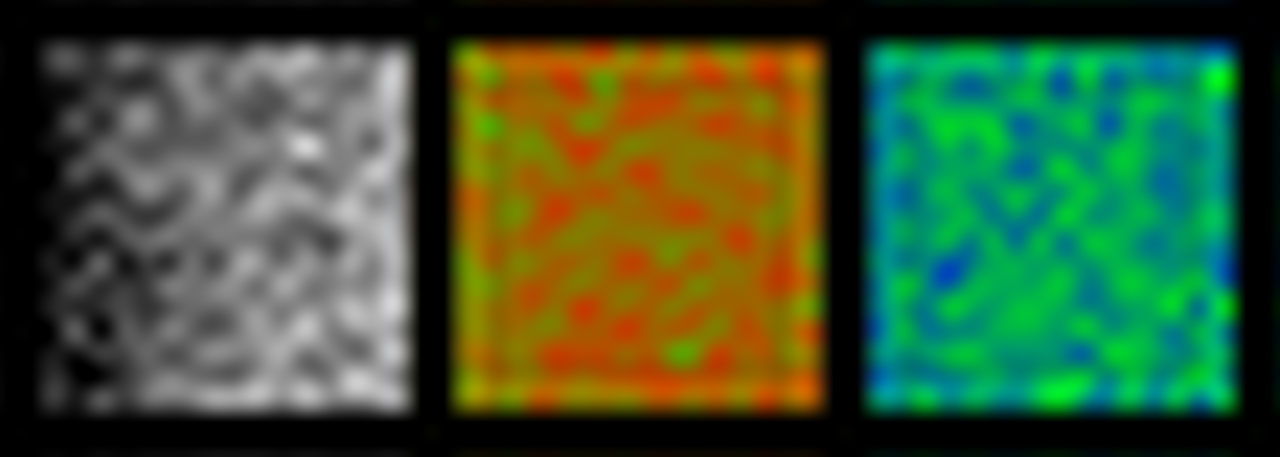}
    \includegraphics[width=0.2\linewidth]{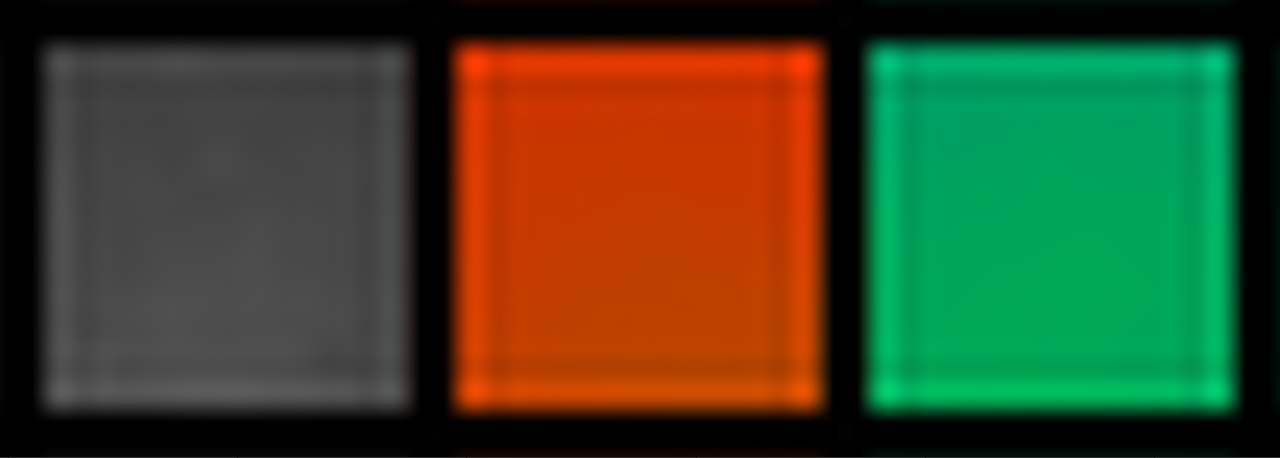}
    \includegraphics[width=0.2\linewidth]{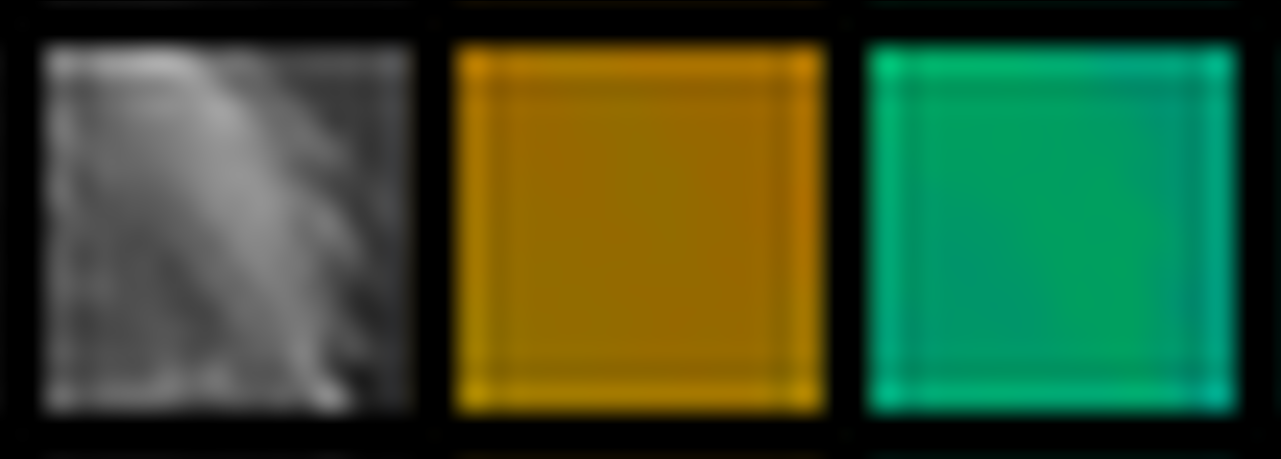}
    \includegraphics[width=0.2\linewidth]{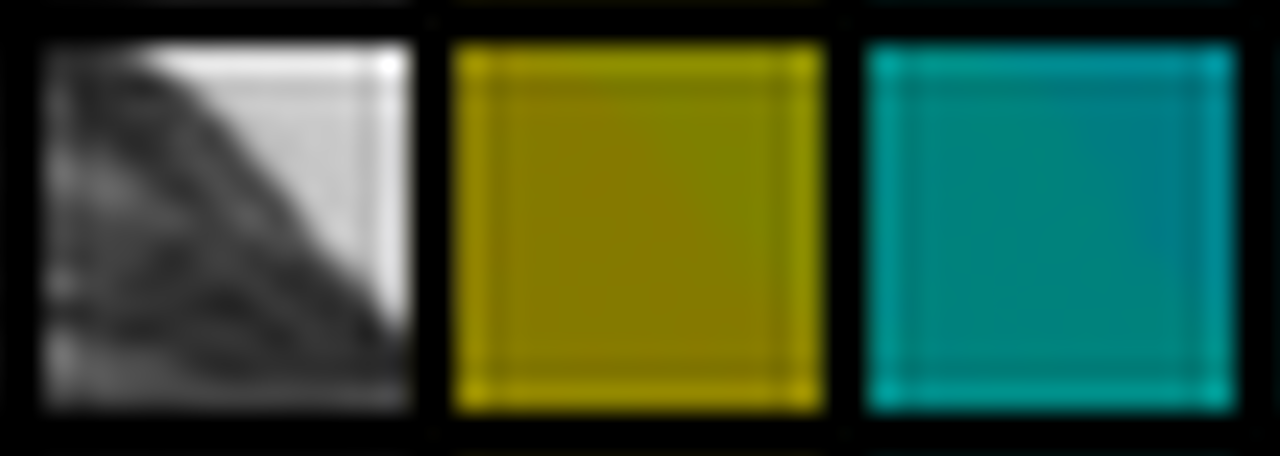}
    \includegraphics[width=0.2\linewidth]{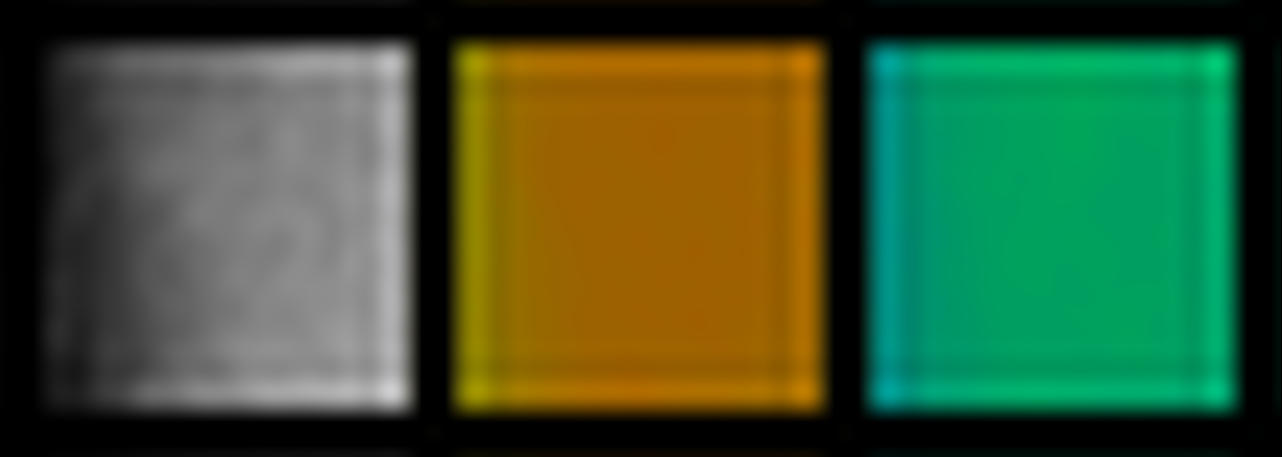}
    \includegraphics[width=0.2\linewidth]{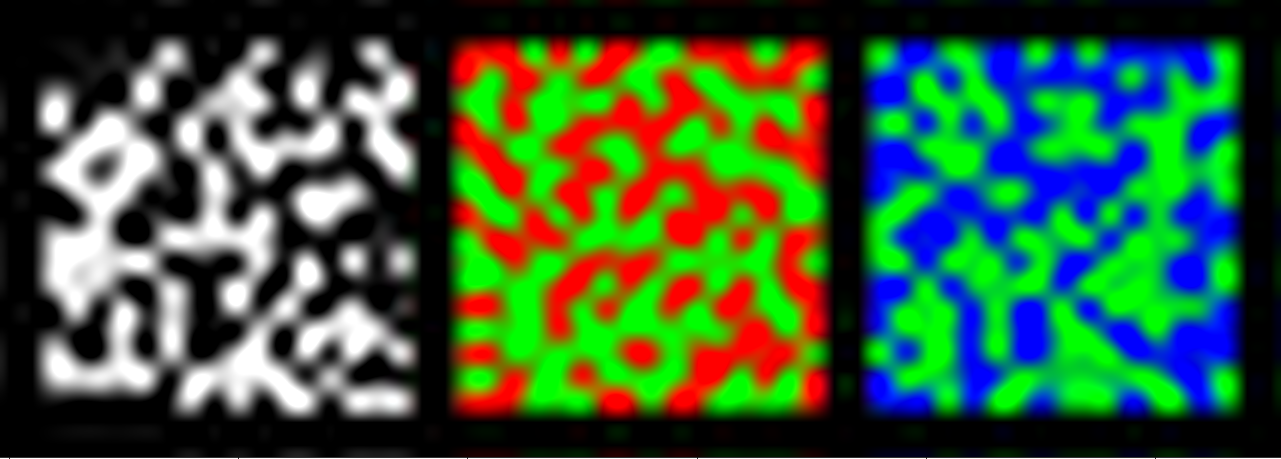}
    \includegraphics[width=0.2\linewidth]{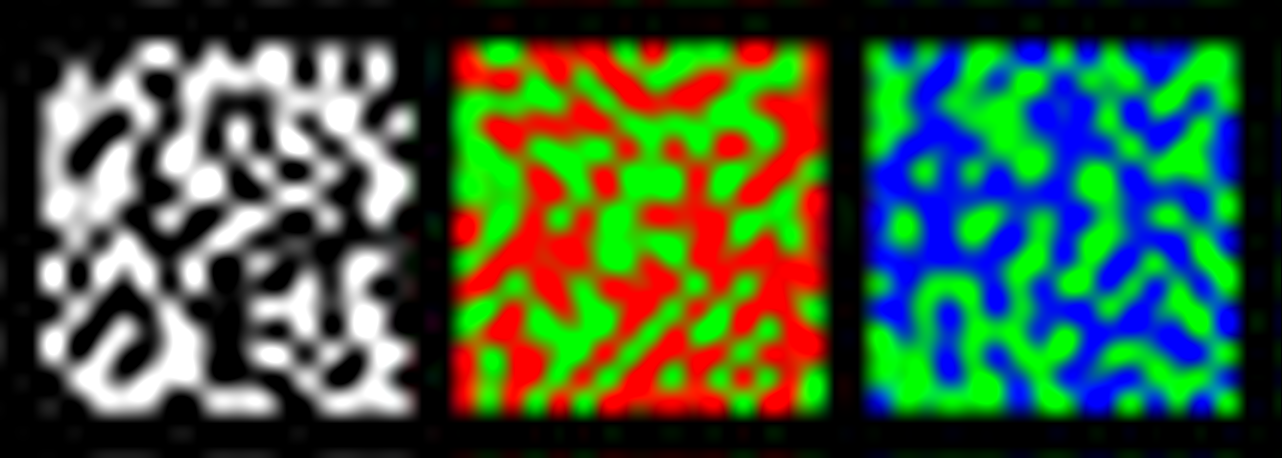}
    \includegraphics[width=0.2\linewidth]{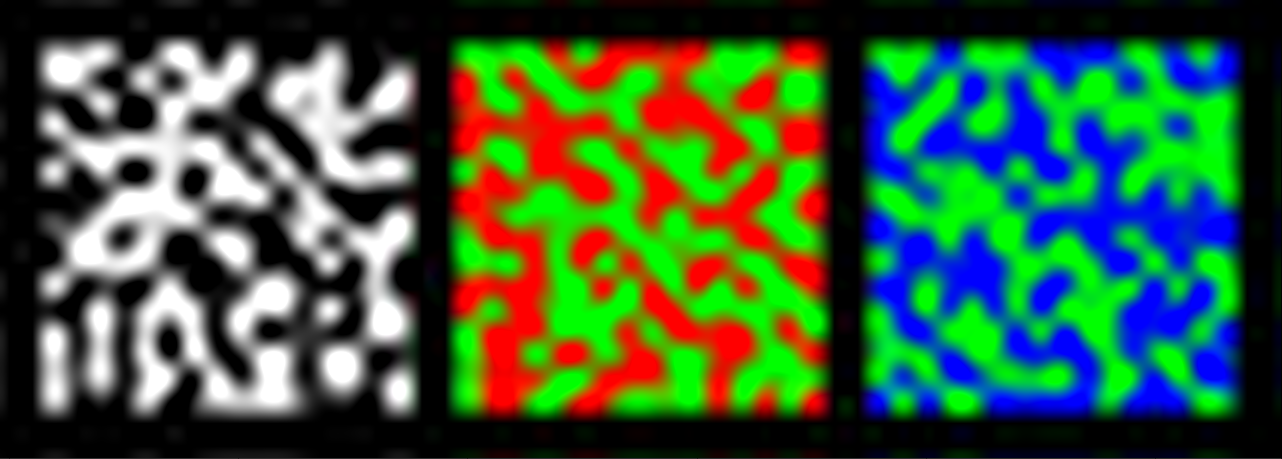}
    \includegraphics[width=0.2\linewidth]{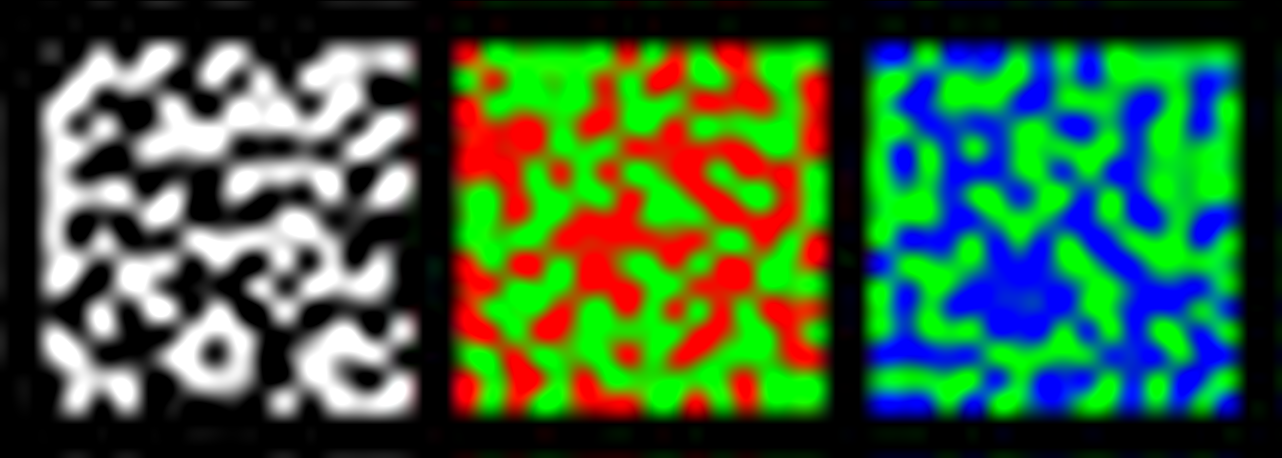}
\end{center}
\caption{Patch-based matching results on YUV spaces: the first row are the noisy patches, the second row are the clean patches that match the noisy patches, and the third is the contrast, representing the noise we are removing. Columns from left to right- three columns as a group- are the images on YUV spaces. For presentation, the patches are scaled to $[0,255]$ for all the channels. This justifies the correctness of our patch transferring scheme within each patch locally. The details are also preserved. Hence, our method is amenable to any size of the image.}
\label{fig:patch_match:yuv}
\end{figure}

We implement the SAC algorithm using Stable-Baseline3\cite{stable-baselines3} from OpenAI\cite{brockman2016openai}. We use OpenAI Gym for setting up the environment. The reward in the environment is computed as stated in Equation \ref{eq:reward} with $b=1.25$, $c=5$, and $PSNR_t = 30.0$. We define the terminating condition $d$ as the following: the SAC will terminate when the actual PSNR reaches the target psnr $PSNR_t$ or the number of epochs reaches the maximum steps allowed, which is 40 in this experiment. The model tries to maximize the reward by optimizing the actions that adjust the trainable weights in the three transformation functions. For minor adjustments, the action space is a set of weight vectors within the bound $(0.999, 1.001)$. The observation space is the actual PSNR score we achieve. In the synthesized CelebA dataset, the learning rate for the model is set to 0.001. The results before and after the recursive self-enhancing procedure are indicated in Table \ref{tab:celeba}, denoted \textit{RSE-RL(before)} and \textit{RSE-RL}, respectively. We also present a set of recursively enhanced images in Figure \ref{fig:recursive}.
\paragraph{Denoising results}
We evaluate our result using PSNR, SSIM\cite{wang2004image}, and UQI\cite{wang2002universal} scores. Some baseline denoising methods, Noise2Void(N2V) and Noise2Noise(N2N) are applied to the synthesized artifacts dataset, and its denoising results are compared with our networks as well. For N2V, the default model for 2D RGB images is trained with 400 noisy images for 50 epochs and tested with 1575 noisy images. In the default model, each image is divided into 128 $16\times 16\times 3$ patches, so a total of 51,200 patches are fed into the network. For N2N, the pre-trained N2N model for Gaussian noise is tested on 1575 noisy images.
\begin{wraptable}{r}{7cm}
\caption{CelebA Results Comparison}
\label{large-table}
\begin{center}
 \begin{tabular}{||c c c c||} 
 \hline
  & PSNR & SSIM & UQI \\ [0.5ex] 
 \hline
 Image with Artifacts & 16.64 & 0.5835 & 0.7327 \\
 
 N2V\cite{krull2019noise2void}& 21.66 & 0.7242 & 0.9249 \\
 
 N2N\cite{moran2020noisier2noise}& 26.60 & / & / \\
 
 VAE& 26.81 & 0.7621 & 0.9604 \\
 
 
 RSE-RL(before) & 28.83 & 0.8322 & 0.9721 \\
 
 RSE-RL & \textbf{29.03} & \textbf{0.8339} & \textbf{0.9731} \\
 
 RSE-RL(S) & 28.79 & 0.8317 & 0.9717 \\[1ex] 
 \hline
\end{tabular}
\end{center}
\label{tab:celeba}
\end{wraptable}
Table \ref{tab:celeba} provides the results obtained using 2250 training images and 11250 testing images. The training images are divided into 1 million $16\times 16\times 3$ patches (441 patches per image) and fed into the network. The result of the baseline method is compared to the results of our RSE-RL. We can observe a significant improvement in all the quality metrics using RSE-RL. A visualization of our denoised image results after self-enhancement can be found in Figure \ref{fig:celeba:results}.
Among the results in Table \ref{tab:celeba}, the RSE-RL achieves the best performance regarding all three metrics. After applying the RL model for self-improvement, we can observe enhancements on all three quality metrics. This demonstrates that our self-enhancing model can be optimized during the testing iterations.
Another observation is that the training data size does not significantly affect RSE-RL's denoising performance. We perform an experiment using 450 training images (0.2 million patches) and the same set of testing images to demonstrate this feature.
We present the result in Table \ref{tab:celeba}, denoted as \textit{RSE-RL(S)}. This result can be compared with \textit{RSE-RL(before)} to observe how the training data size affects the performance. This observation provides an effective way of utilizing this network to largely reduce training time.
\paragraph{YUV Visualization}
The results indicate that learning the transformations of the latent spaces is effective. By comparing the three latent subspaces $Z_Y$, $Z_U$, and $Z_V$, we can observe that the noise has the largest impact on the Y space $Z_Y$, which represents the luminance (brightness) of the image. There is no significant difference between the noisy and clean patch representations on the other two subspaces. This indicates that Gaussian noise significantly impacts brightness compared to chrominance (represented by U and V). The visualized latent subspaces can be seen in Figure \ref{fig:latent}. We also present the reconstructed patches from the YUV subspaces in Figure \ref{fig:patch_match:yuv}

\subsection{SIDD Denoising Result}
\label{sec:sidd}

\begin{figure}[!ht]
\begin{center}
    \includegraphics[width=0.47\linewidth]{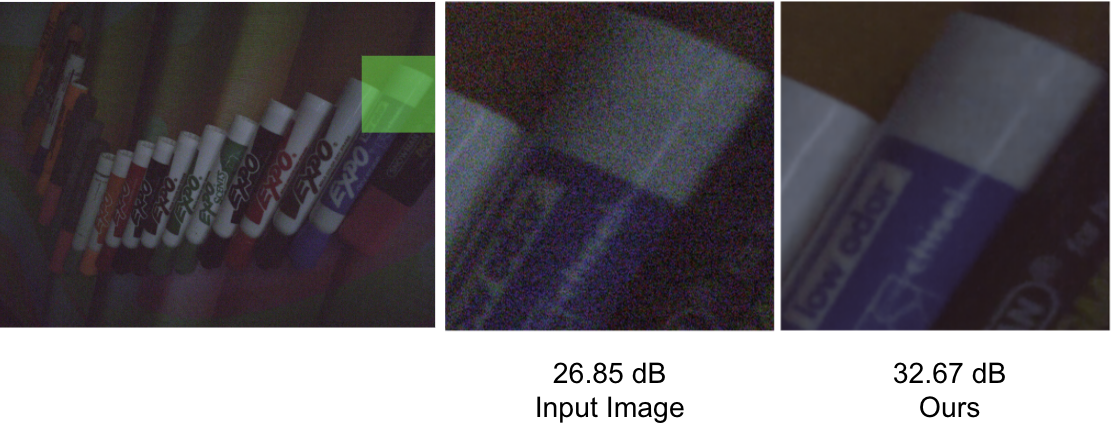}
    \includegraphics[width=0.47\linewidth]{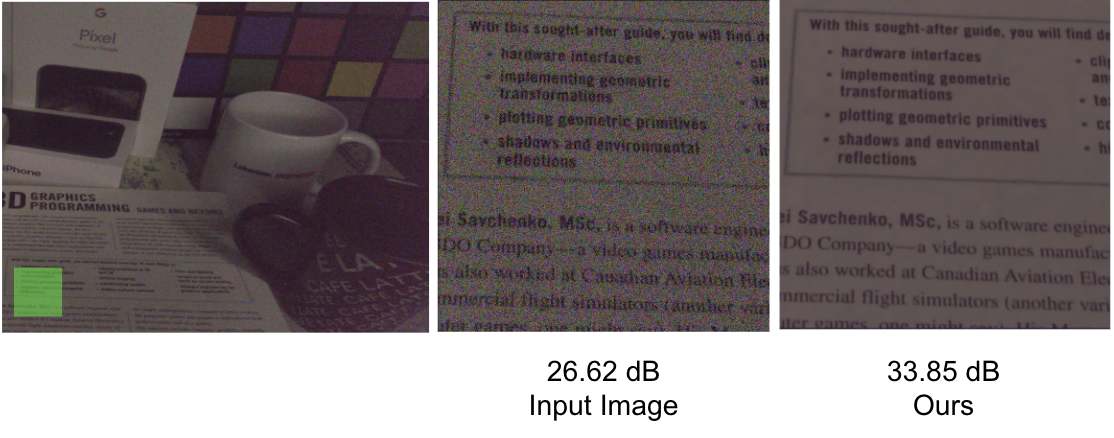}
\end{center}
\caption{SIDD Denoising Result: to better visualize the denoising result, we zoom in at the area covered by the green rectangle in each figure on the left column. The figures from the left column to the right column are: original noisy images from SIDD, zoomed noisy patches, and zoomed denoising patches, respectively.}
\label{fig:sidd}
\end{figure}

\paragraph{Dataset Construction}
Smartphone Image Denoising Dataset (SIDD) is a benchmark for denoising algorithms. It contains about 30,000 noisy images captured by five representative smartphone cameras under 10 lighting conditions and their ground truth. Each noisy image features a mixture of artifacts generated under realistic scenarios. Artifacts caused by ISO levels, illumination, lighting conditions, and signal-dependent noise can all be seen within the dataset.

To test our network, we sample 320 sRGB images as the training data and used the SIDD Benchmark Data, which contains 40 noisy sRGB images and their ground truth, as the testing set. For each benchmark image, we sample 32 patches instead of taking an evaluation measure on the entire image. Besides, we preprocess the training dataset and divide each image, both noisy and growth truth, into $24 \times 24 \times 3$ patches, with 8 pixels overlapping. Consequently, there is a total of 11.19 million patches used for training. 

\begin{table}[t]
\caption{SIDD sRGB to sRGB Results (Small Scale)}
\begin{center}
 \begin{tabular}{||c c c |c c c||} 
 \hline
  & PSNR & SSIM & & PSNR & SSIM\\ [0.5ex] 
 \hline
 Noisy Image & 31.18 & 0.831 & BM3D\cite{dabov2007image}& 25.65 & 0.685\\
 
 NLM\cite{buades2005non}& 26.75 & 0.699 & KSVD\cite{1710377} & 26.88 & 0.842\\
 
 DANet\cite{fu2019dual} & 39.25 & 0.955 & RDB-Net\cite{zhang2018residual} & 38.11 & 0.945\\

 VAE& 31.89 & 0.874 &  &  &\\ 
 
 RSE-RL(before) & 32.38 & \textbf{0.891} & RSE-RL & \textbf{32.53} & 0.887\\ 
 \hline
\end{tabular}
\end{center}
\label{tab:sidd}
\end{table}
\paragraph{Experiment Setup}
Our encoder (with approximately 2.2 million parameters) and decoder (with 1.6 million parameters) structures are the same as the one defined in the Experimental Setup of Section \ref{sec:celeba}. Again, we use a single 12GB NVIDIA Tesla K80 GPU for training and testing on SIDD, and the training batch size is 128. The same optimizer is chosen as we did for the Synthesized Noisy CelebA dataset, defined in Section \ref{sec:celeba}. Our model's parameters were optimized after 20 epochs of training. 
The SAC setting is identical to Section \ref{sec:celeba}, except for we set $PSNR_t = 34.0$ in Equation \ref{eq:reward}. The results before and after self-enhancement are shown in Table \ref{tab:sidd}, denoted as \textit{RSE-RL(before)} and \textit{RSE-RL}, respectively.

\paragraph{Denoising Results}
The results show that our self-enhancing RL model contributes a small enhancement to PSNR, demonstrating that our RL model can improve the denoising results. Since we are only involving PSNR in the reward function, we can only observe some improvements in terms of PSNR.
\begin{figure}[ht]
\begin{center}
    \includegraphics[width=0.7\linewidth]{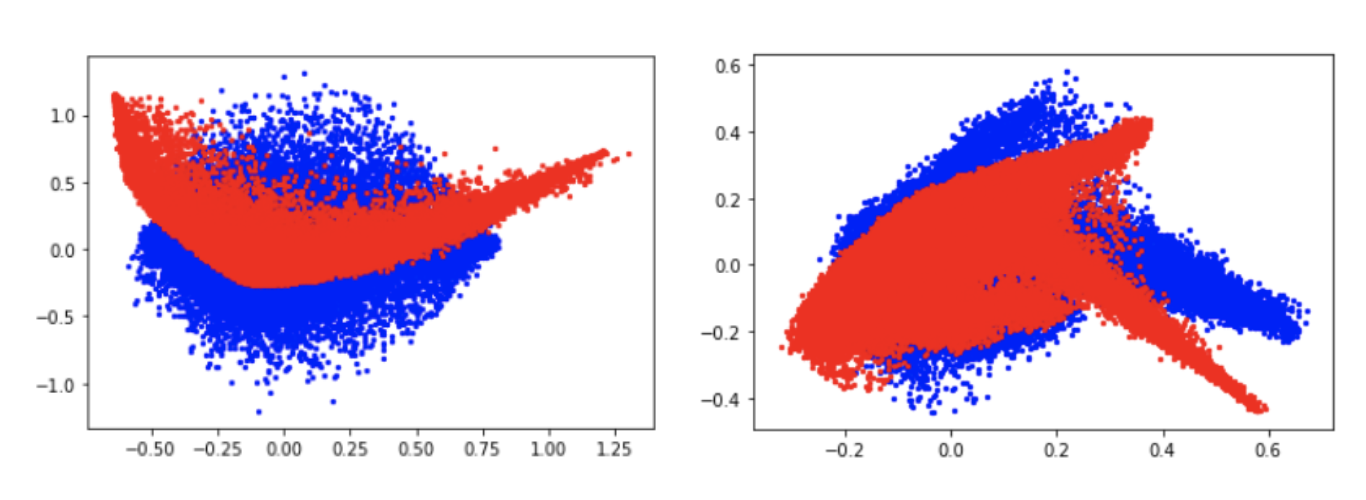}
\end{center}
\caption{Y latent subspaces display: the figure shows the difference between noisy and clean patches our network captured in Y latent subspaces. We show the $Z_Y$ spaces in CelebA(left) and SIDD(right) datasets respectively. Blue points are noise projections, and red points are clean projections. Principle Component Analysis is applied to reduce the dimension of the latent subspaces into 2 for visualization.
}
\label{fig:latent}
\end{figure}
\begin{figure}[ht]
\begin{center}
    \includegraphics[width=0.3\linewidth]{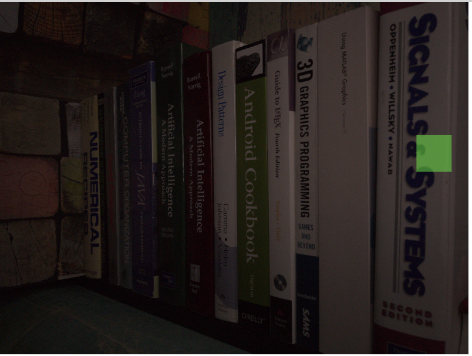}
    \includegraphics[width=0.225\linewidth]{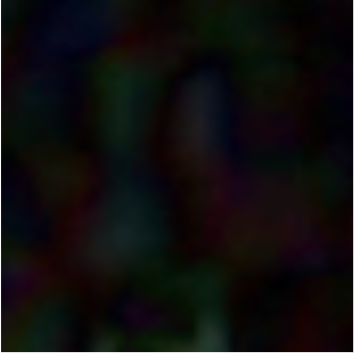}
    \includegraphics[width=0.8\linewidth]{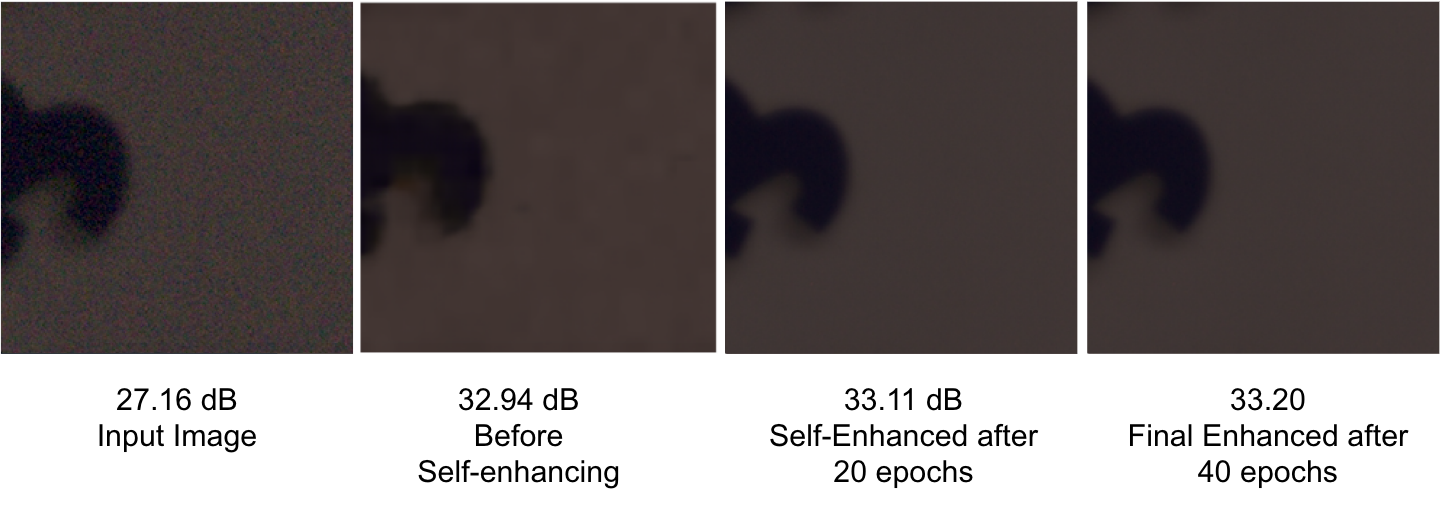}
\end{center}
\caption{Patch enhancement on SIDD: the top left figure shows where we select the patch; the top right figure shows the difference between this patch before and after the self-enhancement. The bottom row shows the patch enhancement over epochs.}
\label{fig:sidd:enhancing}
\end{figure}

\begin{figure}[ht]
\begin{center}
    \includegraphics[width=0.36\linewidth]{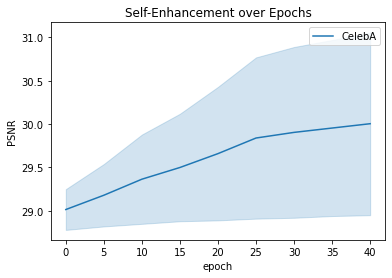}
    \includegraphics[width=0.36\linewidth]{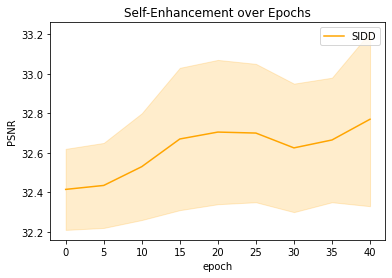}
\end{center}
\caption{Self-enhancement in PSNR over epochs: the left and right figures present the enhancement results on CelebA and SIDD, respectively. We also indicate the variance of our enhancement quality due to the unstable character of the RL algorithm. We perform the RL algorithm multiple times on the same set of images and compute the variances. As the figures presented, the variances of the image qualities tend to increase while the number of epochs increases.}
\label{fig:epoch_psnr}
\end{figure}

Table \ref{tab:sidd} also lists benchmark denoising methods and deep learning methods used to compare against our network. In the table, noisy images are the images before denoising procedures; BM3D, NLM, and KSVD are the benchmark non-DL results; DANet and RDB-Net are two of the state-of-art deep learning methods used for comparison with our method. The performance of our model is significantly better than traditional methods. Figure \ref{fig:sidd} shows the visualized results of self-enhanced denoised images. As for efficiency comparison, our RSE-RL only contains $2.5$ million parameters in total, whereas DANet contains $\sim 60$ million parameters, leading our network train much faster than the state-of-the-art structure.

The images from SIDD consist of the same realistic artifacts generated by smartphone cameras. This leads to the same transformation in the latent space for every patch since each patch consists of the same types of noises. From figure \ref{fig:latent} we can observe a transformation between the noisy patch projections and the clean patch projections on the latent space.

\begin{figure}[!htbp]
\begin{center}
    \includegraphics[width=0.72\linewidth]{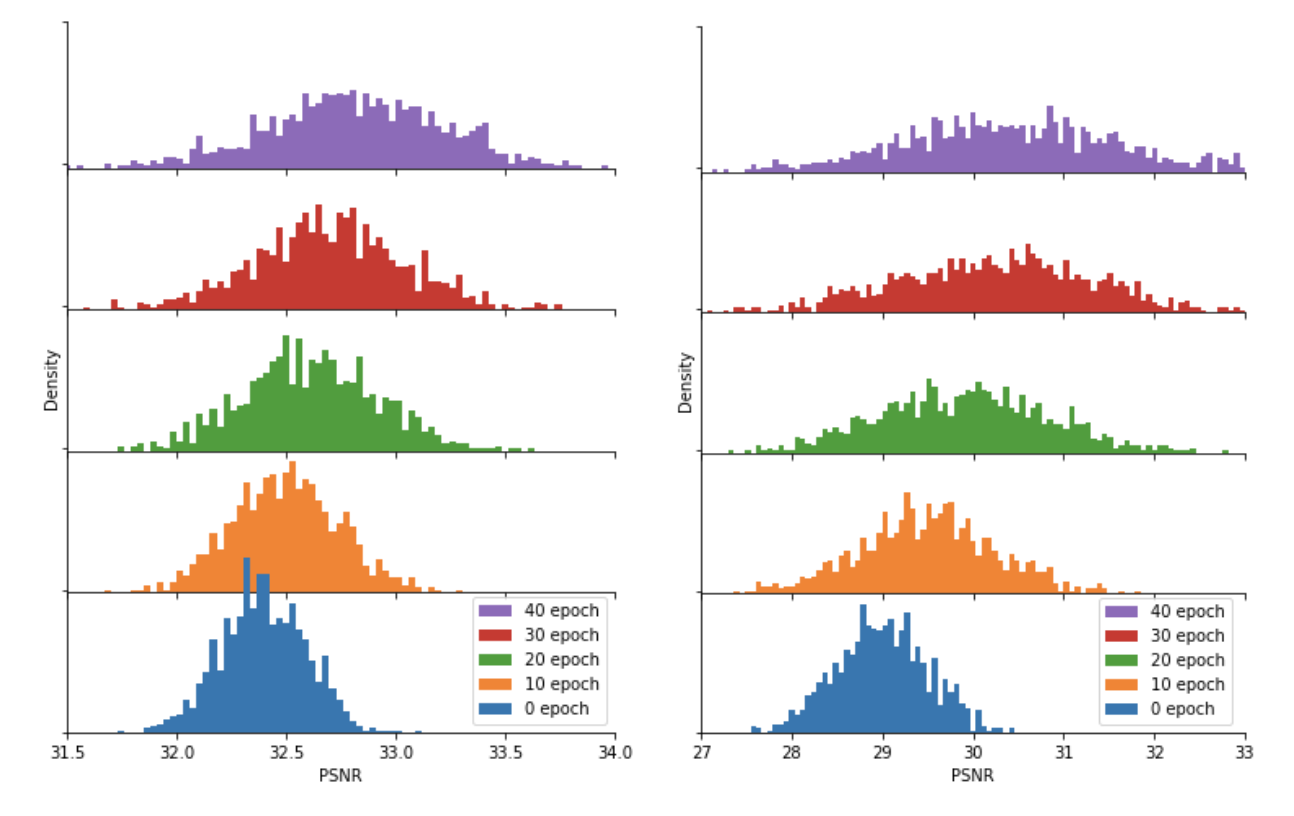}
\end{center}
\caption{Patch-PSNR distribution at each epoch: the left and right figures present the enhancement results on CelebA and SIDD, respectively.}
\label{fig:patch_epoch_psnr}
\end{figure}
\paragraph{Self-Enhancing Visualization}
We then select a patch from an image in SIDD to visualize the self-enhancement on this zoomed region. We show the enhancement visualization of the zoomed region in Figure \ref{fig:sidd:enhancing}. We can observe a small improvement in this patch before and after applying the RL algorithm. To quantify this improvement, we compute the PSNR scores for some sample images and show the relationship between the PSNR and the number of epochs of the RL algorithm in Figure \ref{fig:epoch_psnr}.
We also compute the PSNR score for the patches and present the distribution of patch-PSNR at each epoch of self-enhancing in Figure \ref{fig:patch_epoch_psnr}. We can observe a slight increase in the average PSNR of all the patches. Simultaneously, the variance of the patch-PSNR distribution also grows while the number of epochs increases.

\section{Conclusions}

\label{sec:conclusion}

We have presented a Recursive Self-Enhancing Reinforcement Learning  (RSE-RL) model, for a self-improving camera ISP built upon adaptive and heterogeneous image filtering and patch specific policy learning. The patch-based transformations are  progressively trained in multiple latent subspaces to identify and rectify, spatially heterogeneous and camera specific lens-color acquisitional image artifacts.
We define our action spaces and reward function for a self-enhancement framework and further discuss its potential for real-word Camera ISPs. Nonetheless, our work is an early-stage exploration and exploitation solution. We are moving toward considering optimizing patch scan ordering specific protocols for more efficient processing,  and more complicated environmental settings, to further strengthen our RSE\_RL Camera ISP.

\noindent
{\it Acknowledgements}
 The research was supported in part from the Peter O’Donnell Foundation. Further CB was funded in part from NIH DK129979, and from a grant from the Army Research Office accomplished under Cooperative Agreement Number W911NF-19-2-0333. .

\bibliographystyle{springer_template/bibtex/splncs03}
\newpage
\bibliography{ref}

\newpage
\section*{Appendix}

\subsection*{Detailed Setup of Our RSE-RL Framework}

\paragraph{Architecture of Patch Transformation - Correspondence Network}

Our network architecture transforms the image patches from RGB to YUV channels before encoding. The RGB-YUV transformation is defined as

\begin{equation*}    \begin{bmatrix} P_y\\P_u\\P_v\end{bmatrix} = 
    \begin{bmatrix} 0.299 & 0.587 & 0.114\\-0.147 & -0.289 & 0.436 \\ 0.615 & -0.515 & -1.000 \end{bmatrix}
    \begin{bmatrix}P_r\\P_g\\P_b\end{bmatrix}
\end{equation*}

An encoder $q$ encodes the YUV channels respectively and projects the patch information on three latent subspaces $Z_y$, $Z_u$, and $Z_v$.
The dimension of one subspace is set to 72 for both sets of experiments, hence the latent space dimension is 216. In each of the latent subspaces, both clean and noisy patch representations are projected, and we want to learn a transformation that matches noisy patch representations to clean patch representations. The transformations $T_{y},T_{u},T_{v}$ are defined and operated in their corresponding latent subspaces. Each of the transformation $T_s$ ($s \in \{y,u,v \}$) is a three-layer MLP, with identical dimension layers and ReLU activation. Each transformation is trained to match from a noisy patch representation $z_s^b$ to a clean patch representation $z_s^c$ within its latent subspaces, the loss function is defined in Equation \ref{eq:loss:enc:KL}.

\subsection*{Additional Experimental Results of Our RSE-RL framework}

\paragraph{Decomposed Subspace Visualization}

\begin{figure*}[!htbp]
\begin{center}
    \includegraphics[width=0.85\linewidth]{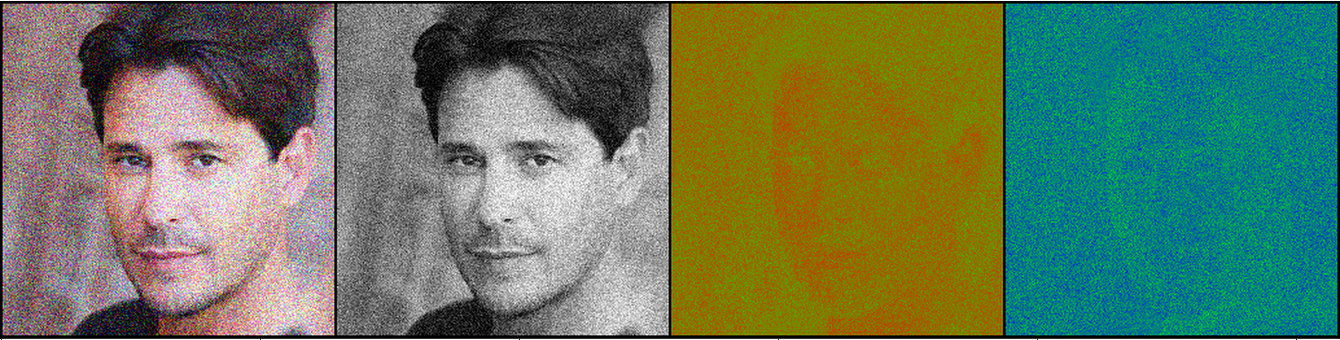}

    \includegraphics[width=0.85\linewidth]{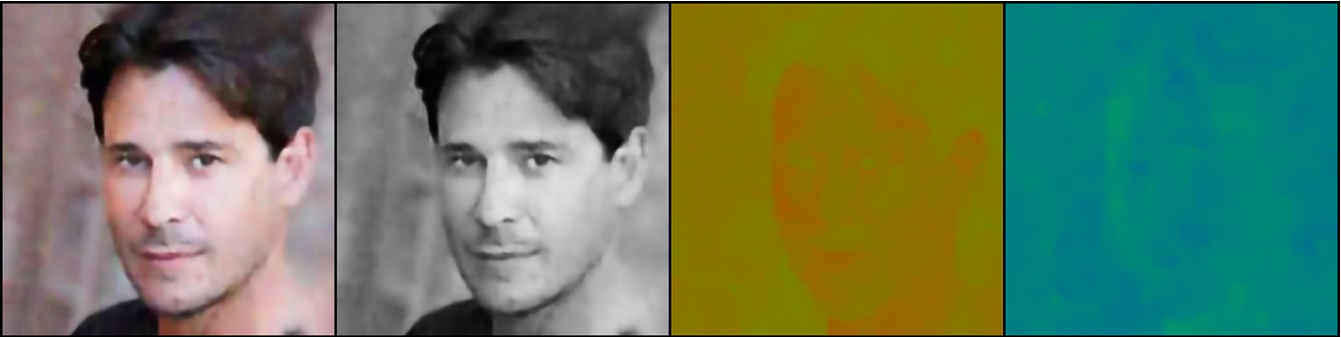}
    \includegraphics[width=0.85\linewidth]{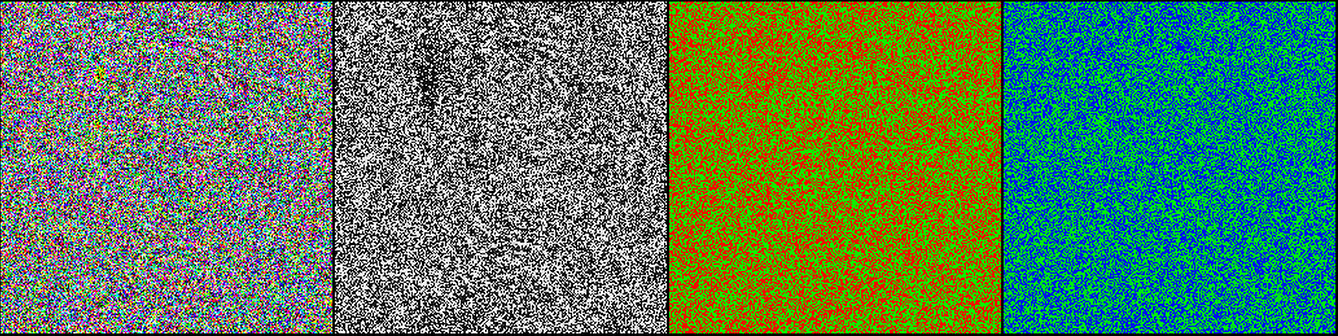}
\end{center}
\caption{CelebA Denoising Visualization in YUV Spaces: images on the top row are images that contain the synthesized artifacts (Gaussian Noise). The images on the second row are the denoising result from our RSE-RL. And the images on the bottom row show the difference between the first two rows, which are the expected noises we removed by the network. The images are scaled to $[0,255]$ for all channels. Columns from left to right show the images on RGB channels, Y space, U space, and V space, respectively. Our method reveals and remove the noise decomposed in three channels respectively.}
\label{fig:denoise:yuv}
\end{figure*}

\begin{figure*}[!htbp]
\begin{center}
    \includegraphics[width=0.9\linewidth]{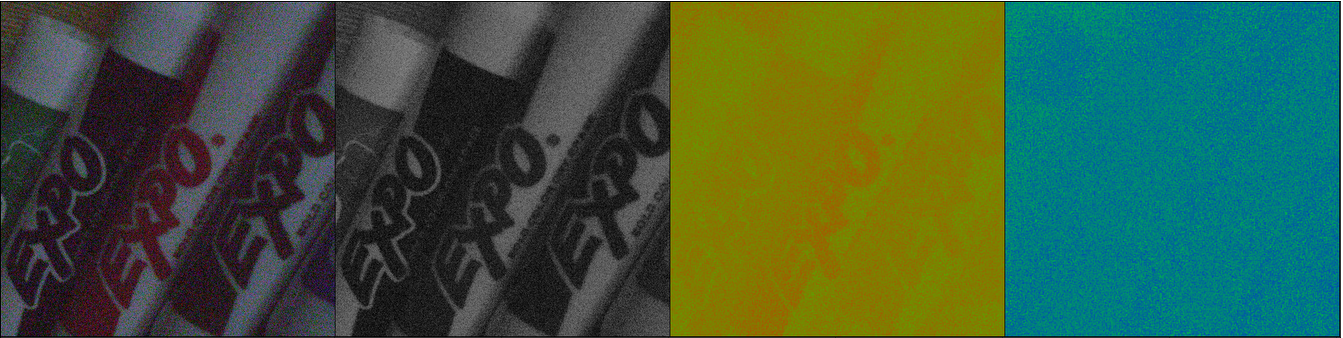}
    \includegraphics[width=0.9\linewidth]{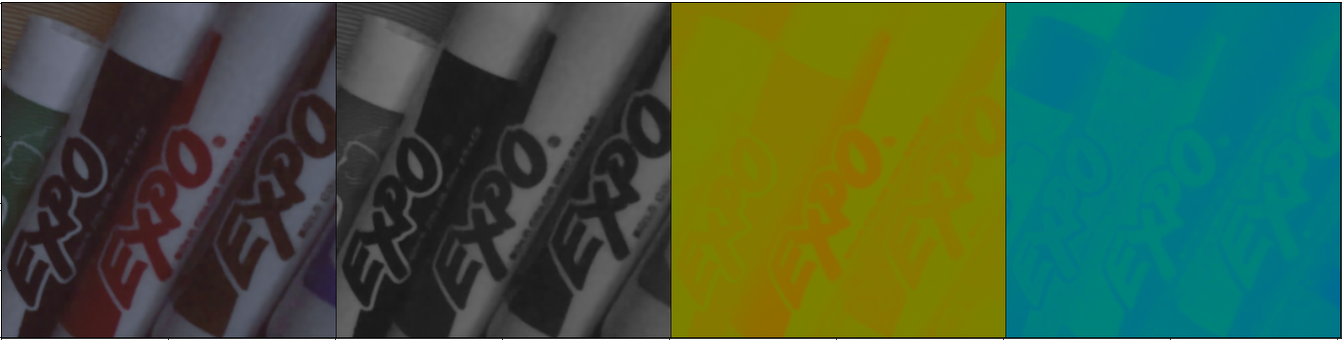}
    \includegraphics[width=0.9\linewidth]{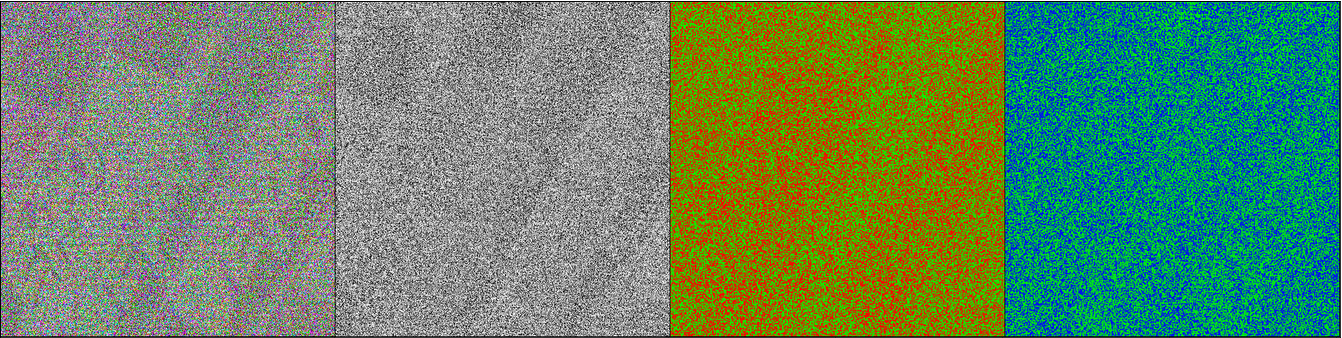}
\end{center}
\caption{SIDD Denoising Visualization in YUV Spaces: specifications are identical to Figure \ref{fig:denoise:yuv}. The figure demonstrates a noise removal over channels and show our patch based method can apply to large-scale, realistic image as well.}
\label{fig:sidd_denoise:yuv}
\end{figure*}

The following sets of figures show the denoising results on each of the Y, U, and V spaces, which further demonstrate how the noises are removed from each space. The figures \ref{fig:denoise:yuv} and \ref{fig:sidd_denoise:yuv} present the examples on both our synthesized CelebA dataset and SIDD dataset. Furthermore, we show some patch-wise matching in Figure \ref{fig:patch_match:yuv}. The figure gives several specific patches as examples for demonstrating how the noisy patches map to the clean patches. It also shows the noises on the patches specifically, and we can observe the noises on the Y, U, and V spaces.

\paragraph{Iterative Image Enhancement Improvement Framework using our RSE-RL network}

We also present how the images are recursively self-enhanced in the reinforcement learning framework in Figure \ref{fig:recursive}. The resulting images over the CelebA dataset have shown us an improvement using the RL backbone. Without RL training, our VAE performance yield at a local minimum while the weight updating under feedback control gives us closer to the optimal result. We would also justify that, based on our observation, random initialization of the entire scheme would yield a much slower convergence rate and an extreme low-PSNR local minimum, and thereby the pre-trained network parameter initialization is crucial for generating high-quality enhanced images.

\paragraph{Justification of Removing Block Artifacts}

There might be the case where our filter generates patch-based enhancement result locally while ignoring the neighboring patches. The one-to-one correspondence from noisy patches to clean patches might cause additional block artifacts, as stated earlier in Section 3. We propose the post processing using overlapping patch smoothing, or additional deblocking algorithm to correct the newly introduced artifacts. Below we show an ablation study under the influence of overlapping patch selections and the use of deblocking artifacts.

\begin{figure*}[!htbp]
\begin{center}
    \includegraphics[width=0.18\linewidth]{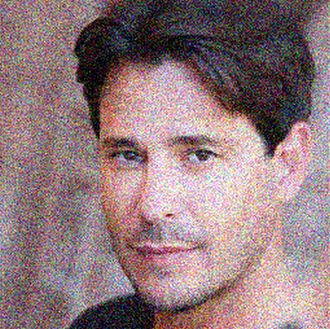}
    \includegraphics[width=0.18\linewidth]{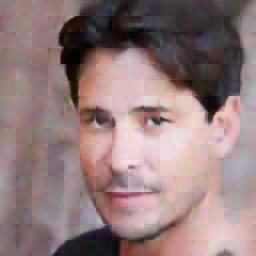}
    \includegraphics[width=0.18\linewidth]{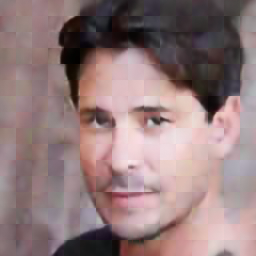}
    \includegraphics[width=0.18\linewidth]{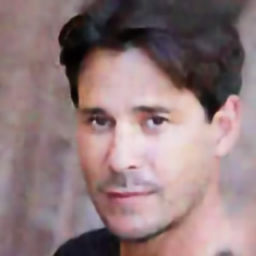}
    \includegraphics[width=0.18\linewidth]{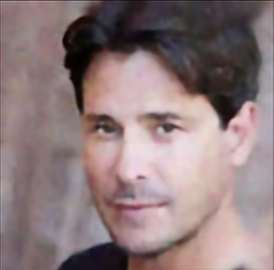}
\end{center}
\caption{Deblocking Results: this figure shows the results of a deblocking method\cite{kim1999deblocking}, as well as our overlapping patch smoothing alternative. It shows that our overlapping patch smoothing method can remove the  block artifacts that may be created by our patch-based scheme. The columns from left to right show the noisy image, image composed by patches without overlapping, non-overlapping patches with deblocking enhancement, image with overlapping patch smoothing, and image with overlapping patch smoothing + deblocking enhancement. The PSNR scores for these images from left to right are 19.89, 29.51, 29.50, 30.31, and 30.31.
}
\label{fig:deblocking}
\end{figure*}

In general testing, we compare the qualities between the non-overlapping patches and overlapping patches, as well as the qualities before and after using the deblocking method \cite{kim1999deblocking}. The average PSNR for images composed of non-overlapping patches is 27.8214. And we can observe obvious blocking artifacts on the edge of the patches (in Figure \ref{fig:deblocking}). When we apply the deblocking method, the average PSNR is slightly reduced to 27.8212 and the block artifacts can still be visualized.

By comparison,  after we apply the overlapping patches, there is a smooth transition on each of the edges between the two blocks. The average PSNR for images composed of overlapping patches is 28.84, which is a significant enhancement. We can also observe the enhancement in the figure \ref{fig:deblocking}. However, we applied the deblocking method to the images with overlapping patches and there is no observable improvement, while the average PSNR stays the same.


\end{document}